\pdfoutput=1

\PassOptionsToPackage{table}{xcolor}
\documentclass[11pt]{article}

\usepackage[final]{acl}

\usepackage{times}
\usepackage{latexsym}
\usepackage{amsmath}
\usepackage{amssymb}
\usepackage{booktabs} 
\usepackage{multirow} 
\usepackage{arydshln}
\usepackage{algorithm}
\usepackage{algpseudocode}
\usepackage{tabularx}   
\usepackage[T1]{fontenc}

\usepackage[utf8]{inputenc}

\usepackage{microtype}

\usepackage{inconsolata}

\usepackage{graphicx}

\usepackage{xspace}

\newcommand{\impliret}{\textsc{ImpliRet}\xspace}

\newcommand\blfootnote[1]{%
  \begingroup
  \renewcommand\thefootnote{}\footnote{#1}%
  \addtocounter{footnote}{-1}%
  \endgroup
}

\title{\impliret: Benchmarking the Implicit Fact Retrieval Challenge}

\author{
 \textbf{Zeinab Sadat Taghavi\textsuperscript{*}},
 \textbf{Ali Modarressi\textsuperscript{*}},
 \textbf{Yunpu Ma},
 \textbf{Hinrich Schütze}
\\
Center for Information and Language Processing, LMU Munich\\
Munich Center for Machine Learning (MCML)
\\
 \small{
   \textbf{Correspondence:} [\href{zeinabtaghavi@cis.lmu.de}{zeinabtaghavi}, \href{amodaresi@cis.lmu.de}{amodaresi}]@cis.lmu.de
 }
}

\newcounter{notecounter}
\newcommand{\enotesoff}{\long\gdef\enote##1##2{}}

\enotesoff

\long\def\devour#1{}

\begin{document}
\maketitle
\begin{abstract}
Retrieval systems are central to many NLP pipelines, but often rely on surface-level cues such as keyword overlap and lexical semantic similarity. To evaluate retrieval beyond these shallow signals, recent benchmarks introduce reasoning-heavy queries; however, they primarily shift the burden to query-side processing techniques -- like prompting or multi-hop retrieval -- that can help resolve complexity.  In contrast, we present \impliret, a benchmark that shifts the reasoning challenge to document-side processing: The queries are simple, but relevance depends on facts stated implicitly in documents through temporal (e.g., resolving “two days ago”), arithmetic, and world knowledge relationships.  We evaluate a range of sparse and dense retrievers, all of which struggle in this setting: the best nDCG@10 is only 14.91\%. We also test whether long-context models can overcome this limitation. But even with a short context of only thirty documents, including the positive document, GPT-o4-mini scores only 55.54\%, showing that document-side reasoning remains a challenge. Our codes are available at \href{https://github.com/ZeinabTaghavi/IMPLIRET}{github.com/ZeinabTaghavi/IMPLIRET}.\blfootnote{\textsuperscript{*} Equal Contribution.}

\end{abstract}

\section{Introduction}
Retrieval systems play a pivotal role in many NLP applications, enabling models to utilize relevant information from large corpora such as document collections, web pages, or conversational histories \cite{Lewis2020RAG, gao2023retrieval}.
Relevance in retrieval can be established through a range of connections, from explicit lexical or semantic similarity to more implicit, context-dependent associations.
However, widely used retrieval systems are highly reliant on surface-level cues such as exact matches, repetition, or where a fact appears in the text \cite{ram-etal-2023-token, coelho-etal-2024-dwell, fayyaz2025collapse}. 
Additionally, many popular benchmarks (e.g., BEIR
\citep{thakur2021beir}) do not surface these issues as their
queries have lexical overlap with relevant documents \cite{shao2025reasonir}.
There are attempts to create reasoning-intensive datasets that push beyond lexical and surface-level matches. 
For instance, RAR-b \cite{xiao2024rar} reframes multiple-choice reasoning tasks into retrieval problems, BIRCO \cite{wang2024birco} collects multi-faceted questions across five domains, and BRIGHT \cite{su2025bright} uses full StackExchange problem descriptions as queries against the pages they cite. 
Since the reasoning burden lies on the query side,
techniques like query expansion, chain-of-retrieval inference, or agentic retrieval can help models handle complex prompts and outperform standard retrievers \cite{wang2025chainofretrieval, song2025r1searcher, li2025searcho1agentic}.
\begin{figure}[t]
	\centering
	\includegraphics[width=0.9\linewidth]{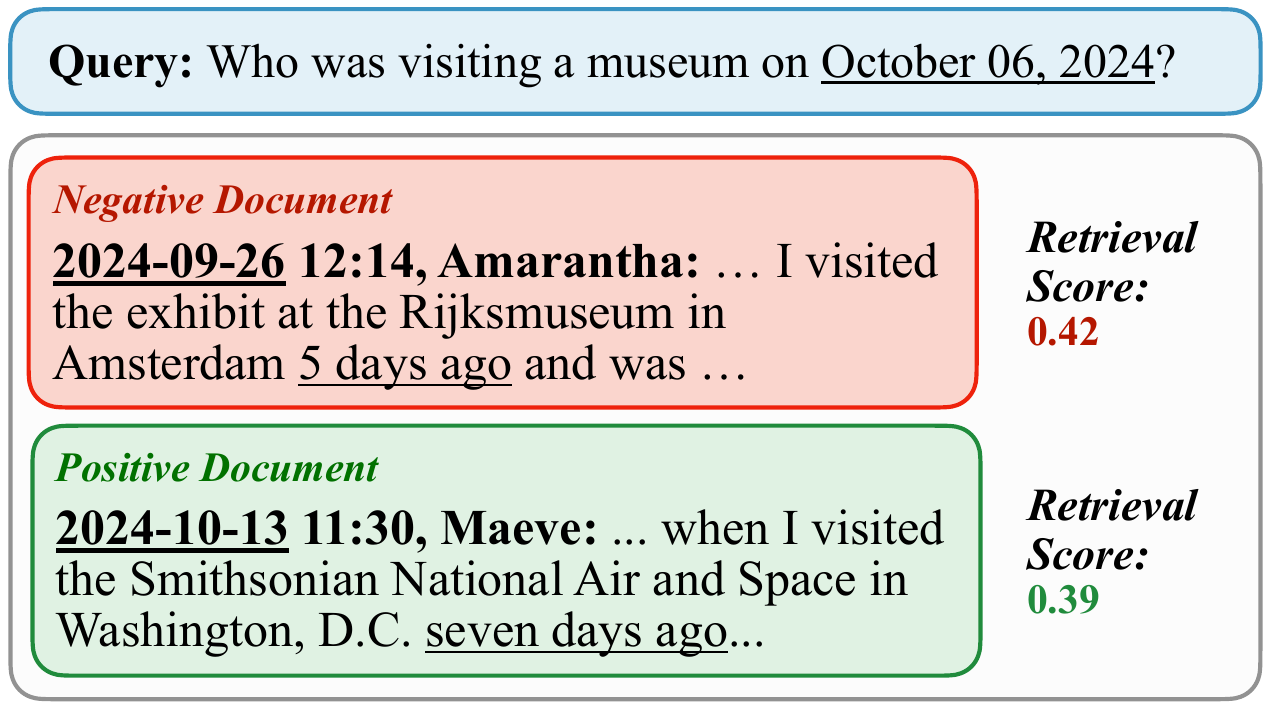}
	\caption{\textbf{An example from \impliret:} a query
            and two sample documents, negative and positive.
            Retrieval of the relevant positive document requires
            surfacing implicit knowledge: that
Maeve visited the Smithsonian on 2024-10-06.}
	\label{fig:MainExample}
\end{figure}

In contrast, we present \textbf{\impliret}, a benchmark that
shifts reasoning to document-side processing: the queries are simple, but relevance depends on facts stated implicitly within the documents, spanning \textbf{arithmetic}, \textbf{temporal}, and \textbf{world knowledge} relationships that require inference to uncover.
Figure \ref{fig:MainExample} gives an example: the correct document requires resolving a reference to a date that is implicit, i.e., not stated directly.
An effective retrieval system must infer such implicit facts from the document content, ideally as part of the indexing process, in order to retrieve the correct result at query time. Yet current retrieval methods fail to capture the implicit signals needed for accurate retrieval. We evaluate sparse and dense approaches, including BM25 \cite{INR-019-BM25}, ColBERT \cite{santhanam-etal-2022-colbertv2}, and Dragon+ \cite{lin-etal-2023-train-dragon}, and observe consistently poor performance:  the best nDCG@10 is only 14.91\% across our benchmark.
To test whether long-context capabilities could mitigate the problem, we evaluate models in a setting where the positive document is included among several distractors. 
While GPT-o4-mini answers correctly when given only the positive document, its performance drops sharply even with just thirty documents in-context, achieving a ROUGE-1 recall of 55.54\%.
Our dataset \impliret introduces a new setting that requires
document-side
reasoning for retrieval  rather
than query-side reasoning. \impliret presents challenges for both retrieval and long-context processing, highlighting the need for models that can reason over implicit information embedded in large corpora.

\section{\impliret}
\label{sec:dataset_generation}

In \impliret, we construct examples whose relevance depends on information that is
implicitly stated in the document, i.e., it can only be discovered through reasoning,
not by surface-level overlap.  \impliret covers three reasoning categories:
\textbf{World Knowledge}, \textbf{Arithmetic}, and \textbf{Temporal}.

We compile a \textbf{collection of implicit-tuple sets}.
Within each set, a tuple links an implicit surface form that appears in a
document to the explicit form that will appear in the query; see
Fig.~\ref{fig:MainExample}, e.g.\
(\textit{``2024-10-13 … seven days ago''}, \textit{``October 06, 2024''}).

For every reasoning category, we create \(N\) such tuple sets.
Each set \(T_i\) (\(i=1,\dots,N\)) contains \(M\) unique tuples (\(|T_i|=M\)).
Tuples in the tuple sets are unique but not guaranteed to be unique throughout the collection of tuple sets. 
Hence, before document generation, we inject distinct auxiliary lexical entities (e.g.\ named entities, speaker names) into each tuple so that the documents
generated from \(T_i\) remain distinguishable from those of \(T_j\) when
\(i \neq j\) (see Appendix~\ref{sec:app_synthesizing_the_context}).

From each tuple in the tuple set, we generate a document,
yielding a pool of documents \(\mathcal{D}_{T_i}\) with \(|\mathcal{D}_{T_i}|=M\).
The document derived from \(t_i \in T_i\) is the \emph{only} positive for the query
constructed from \(t_i\),
whereas all other documents in the global collection
\(
\mathcal{D}= \bigcup_{i=1}^{N} \mathcal{D}_{T_i}
\)
-- including those from tuples \(t'_i \neq t_i\) in the same set and every document
from any other set \(T_j\neq T_i\) -- are treated as negatives.

For each reasoning category, we generate two collections of tuple sets, one realized in the \emph{uni-speaker} style and the other in the \emph{multi-speaker} style, keeping their respective document pools separate to foster surface diversity.
Thus, every query has exactly one positive document, while \emph{every other document in the global collection} serves as a semantically irrelevant negative.
In the remainder of this section, we detail the construction of the implicit-tuple sets and our procedure for generating documents and queries.

\subsection{Generating Tuple Sets}
\paragraph{Arithmetic.}
\label{sec:ds_gen_a}
An arithmetic relation requires simple numerical reasoning. 
For instance, the query ``Which bag costs \$1,600?'' can be answered by ``The Prada bag costs \$2,000, the Gucci bag is 20\% cheaper,'' since \$2,000 × 0.8 = \$1600. 
Here, the model must identify the reference price, interpret the relative statement (``20\% cheaper''), and perform the corresponding computation to infer the answer.
Therefore, each tuple in the implicit tuple set takes the form \(\bigl((p_{1},r,e),p_{2}\bigr)\), where \(p_{1}\) is the base price, \(r\) is the relative multiplier, \(e\in\{\texttt{``Lower''},\texttt{``Higher''}\}\) indicates the direction of the change, and \(p_{2}\) is the queried price (e.g., \(((2000,0.2,\texttt{Lower}),1600)\)).
We apply constraints to ensure that queried prices are unique, realistic, and well-distributed across the tuple set.
Tuples are generated using a sampling algorithm that selects base prices and checks
constraint satisfaction, backtracking as needed until $M$ valid tuples are found
(where $M$ is the target number of documents indicated as
``Docs'' in Table \ref{tab:numerical_report}). Full constraint details and sampling logic are provided in Appendix~\ref{sec:app_ds_gen_a}.

\paragraph{World Knowledge.}
A world knowledge relation connects a textual mention to an external fact.
For instance, the query ``Who was in the UK?'' can be
answered by ``Lenna was at Big Ben,'' based on the implicit fact that Big Ben is located in the UK.
The model must identify the mentioned entity, retrieve the associated world fact, and use it to resolve the query.
Each tuple is encoded as \((\textit{landmark},\textit{country})\), e.g., \((\text{``Big Ben''},\text{``UK''})\).
To build the tuple set, we collect landmark-country pairs that are unambiguous, globally unique, free of lexical cues revealing the country, and refer to specific rather than generic locations.
Candidates are sourced from Wikidata \cite{vrandevcic2014wikidata} and filtered using LLMs, embedding similarity, and web search verification.
Full filtering criteria, prompts, and implementation details are provided in Appendix~\ref{sec:ds_gen_s}. Here, we again generate a set of $M$ tuples of each implicit tuple set.

\paragraph{Temporal.}
A temporal relation involves reasoning over relative dates; we gave an example in Figure
\ref{fig:MainExample}.
The model must identify the reference date (2024-10-13), interpret the relative time expression (``seven days ago''), and compute the resulting absolute date (``2024-10-06''). Each example is represented as a tuple \(\bigl((d_{B},R),D_{L}\bigr)\), where $d_B$ is the base date explicitly mentioned in the document, $R$ is a list of relative offsets (e.g. [``1 day after'', ``2 days after'']), and $D_L$ is the list of resolved explicit dates (e.g., [``March 6th'', ``March 7th'']).
We generate $M$ such tuples under constraints that ensure date uniqueness, broad coverage across a fixed window, and realistic time offsets. Target date sequences are first sampled, then anchored to a base date to define relative expressions. The sampling algorithm verifies constraints and backtracks as needed until a valid set is found.
Further details on constraints and sampling logic are provided in Appendix~\ref{sec:ds_gen_t}.

\begin{table}[pt]
\centering
\scriptsize
\resizebox{\columnwidth}{!}{%
\begin{tabular}{llcrr}
\toprule
\multirow{2}{*}{\textbf{Reasoning}} &
\multirow{2}{*}{\textbf{Style}} &
\multirow{2}{*}{\textbf{Docs}} & \multicolumn{2}{c}{\textbf{Tokens}}\\
\cmidrule(lr){4-5}
&  &  & \multicolumn{1}{c}{\textbf{Avg.}} & \multicolumn{1}{c}{\textbf{Total}}  \\
\midrule
\multirow{2}{*}{Arithmetic} & Uni-speaker    & 1500 & 553 & 830,098 \\
                             & Multi-speaker  & 1500 & 142 & 213,750 \\
\midrule
\multirow{1}{*}{World-}   & Uni-speaker    & 1500 & 471 & 707,047 \\
\multirow{1}{*}{Knowledge}& Multi-speaker  & 1500 & 168 & 253,337 \\
\midrule
\multirow{2}{*}{Temporal}   & Uni-speaker    & 1500 & 479 & 719,226 \\
                             & Multi-speaker  & 1500 & 141 & 212,502 \\
\bottomrule
\end{tabular}%
}
\caption{\textbf{\impliret statistics.}  For each reasoning category and discourse style (uni-speaker vs.\ multi-speaker), we list the number of documents (\textit{50 tuple sets $\times$ 30 docs = 1500}), the average document length, and the total token count.  Every document has exactly one associated query, so the document and query counts coincide.}
\label{tab:numerical_report}
\end{table}

\subsection{Document-Query Pairs}
We generate a document-query pair from every fact tuple, realizing it in one of two styles: 
\emph{uni-speaker} (multi-turn chat) or \emph{multi-speaker} (forum thread).

\paragraph{Uni-speaker (multi-turn chat).}
For each tuple, we create a short multi-turn dialogue.
The same main conversant (e.g., \ ``Alex'') appears in every dialogue within a tuple set
and never appears in any other tuple sets.
To keep the interactions natural, the second conversant’s name changes from one dialogue to the next.
Depending on the reasoning category, the main conversant states which product they bought at
a certain price (Arithmetic), mentions visiting a landmark (World Knowledge),
or describes an activity that occurred on a specific date (Temporal).
The query then targets the implicit fact contained in that statement: the product, person, or
activity linked to the given price, country, or date.

\paragraph{Multi-speaker (forum thread, one post per user).}
Each tuple set receives a single \textbf{prompt} that serves as the thread’s opening post.
For that tuple set, we create a forum thread in which each post is authored by a different user,
realizing one tuple, and all posts respond to the shared prompt.
Thus, the thread mimics a discussion in which several users independently mention
their purchase, visit, or scheduled activity, respectively.
While the underlying actions mirror the uni-speaker setting, the query perspective shifts: instead of asking about an attribute of a known entity, it now asks which entity
(product, person, or activity) satisfies a stated condition such as a price, location, or date.

\paragraph{Generation Pipeline.} In both styles, i.e., in each conversation and post, every message includes a timestamp and speaker name (see Figure~\ref{fig:MainExample}).
In both styles, each example is produced via a three-step pipeline:
(1) Entity binding: We assign entities (e.g., names, items, activities) to each tuple to create a plausible scenario and define the query target;
(2) Document generation: We prompt an LLM to generate a chat
or forum passage that embeds the entity and the implicit part of the tuple, without stating the explicit fact;
(3) Verification: a second model attempts to extract the original tuple;  we retain only examples where the intended fact is fully recoverable.
This pipeline is supported by auxiliary lexical resources, including random names, brand-item pairs, and activity lists, as well as per-reasoning category prompt templates. We use \textsc{Gemma-3-27B-it} \cite{gemma_2025} to synthesize the documents for each tuple.\footnote{Details such as prompts and query templates are available in Appendix~\ref{sec:app_synthesizing_the_context}.} Table \ref{tab:numerical_report} presents \impliret statistics\footnote{The tokens are counted using GPT-2 tokenizer \cite{radford2019language}.}.

\paragraph{Fluency and implicitness sanity check.}
We drew a stratified random sample of 72 instances (query–document pairs; 3 reasoning categories × 2 discourse styles × 12 per cell) and manually assessed each for (i) fluency, (ii) implicit support for the queried fact, and (iii) absence of explicit leakage (i.e., a verbatim statement of the fact). In this sample, all documents were fluent and supported the queries implicitly; under our rubric, we observed no cases of explicit leakage. Further details are provided in  Appendix~\ref{sec:app_human_eval}.

\section{Experiments}

We employ \impliret to probe whether state-of-the-art retrievers can perform \emph{document-side reasoning}. 
Relevant documents are retrieved for each query among those documents that are in its corresponding (reasoning category and discourse style) group.

At test time, each query is compared to all its discourse style documents. Our evaluation covers a wide variety of retrieval methods: sparse lexical baseline \textsc{BM25} \cite{INR-019-BM25,lu2024bm25s}; dense encoders \textsc{Contriever}, \textsc{Dragon+}, and \textsc{ReasonIR} \cite{izacard2021contriever,lin-etal-2023-train-dragon,shao2025reasonir}; late interaction model \textsc{ColBERT v2} \cite{santhanam-etal-2022-colbertv2}; and knowledge graph augmented retriever \textsc{HippoRAG 2} \cite{gutierrez2025rag}. Effectiveness is reported as nDCG@\(k\) in the main text; MRR@\(k\) appears in Appendix~\ref{sec:app_results}.

\begin{table}[t]
\centering
\small
\resizebox{\columnwidth}{!}{%
\begin{tabular}{lcccc}
\toprule
\multirow{2}{*}{\textbf{Retriever}} 
& \multicolumn{3}{c}{\textbf{Reasoning}} & \multirow{2}{*}{\textbf{Average}} \\
\cmidrule(lr){2-4}
& \textbf{W. Know.}
& \textbf{Arithmetic}
& \textbf{Temporal}
& \\

\midrule
\multicolumn{5}{c}{\cellcolor{gray!15} Sparse}\\
\addlinespace
BM25
& 14.69 & 11.06 & 10.98 & 12.24  \\
\specialrule{0.4pt}{\aboverulesep}{\belowrulesep}

\multicolumn{5}{c}{\cellcolor{gray!15} Late-Interaction}\\
\addlinespace
ColBERT v2    
& 15.79 & \textbf{14.96} & 11.99 & 14.25 \\
\specialrule{0.4pt}{\aboverulesep}{\belowrulesep}

\multicolumn{5}{c}{\cellcolor{gray!15} Dense Encoders}\\
\addlinespace
Contriever 
& 16.50 & 13.70 & 12.73 & 14.31 \\
Dragon+ 
& 17.46 & 14.61 & 12.66 & \textbf{14.91} \\
ReasonIR   
& \textbf{18.88} & 10.78 & 11.25 & 13.64 \\
\specialrule{0.4pt}{\aboverulesep}{\belowrulesep}

\multicolumn{5}{c}{\cellcolor{gray!15} Knowledge-Graph Augmented Indexer}\\
\addlinespace
HippoRAG 2
& 16.62 & 14.13 & \textbf{12.83} & 14.53 \\
\bottomrule
\end{tabular}%
}
\caption{\textbf{Retrieval evaluation.} \textbf{nDCG@10} for
  our
  reasoning categories (world knowledge (W. Know.), arithmetic, and temporal),
  averaged over Uni-speaker and Multi-speaker documents) and
  ``Average'' of reasoning.}
\label{tab:ndcg_metrics}
\end{table}

\section{Results}
\label{sec:results}

The nDCG@10 results across all reasoning categories are presented in Table~\ref{tab:ndcg_metrics}.  
The highest average score, 14.91 (achieved by \textsc{Dragon+}), shows the difficulty retrieval models face when reasoning over implicit facts in documents.  
More efficient baselines such as \textsc{Contriever} and \textsc{BM25} perform substantially worse; notably, \textsc{BM25} reaches just 12.24 due to its reliance on surface-level lexical overlap.

Performance varies across reasoning types: the Arithmetic category exhibits the largest performance spread
(14.96 vs.\ 10.78), while it is narrowest
for Temporal 
(12.83 vs.\ 10.98).  
Discourse style also plays a role: \textsc{Dragon+} scores 16.45\% on multi-speaker examples compared to 13.37 on uni-speaker ones, suggesting that stylistic structure affects retrieval difficulty.\footnote{Full results per category and style in Appendix~\ref{sec:app_results}.}

\begin{table}[t]
\centering
\scriptsize
\resizebox{\columnwidth}{!}{%
\begin{tabular}{l|l|ccc|c}
\toprule
\multirow{2}{*}{\textbf{Experiment}}
& \multirow{2}{*}{\(\boldsymbol{k}\)}
& \multicolumn{3}{c|}{\textbf{Reasoning}}
& \multirow{2}{*}{\textbf{Average} } \\
\cmidrule(lr){3-5}
&
& \textbf{W. Know.}
& \textbf{Arithmetic}
& \textbf{Temporal}
& \\
\midrule
\multirow{3}{*}{Llama 3.3 70B}
 & 1   & \textbf{73.79} & \textbf{90.13} & \textbf{81.85} & \textbf{81.92} \\
 & 10  &  27.37 & 16.98 & 25.23 & 23.19 \\
 & 30 & 17.43 & 4.42 & 10.29 & 10.71 \\
\specialrule{0.4pt}{\aboverulesep}{\belowrulesep}
\multirow{3}{*}{GPT-4.1}
 & 1   & \textbf{93.24} & \textbf{92.12} & \textbf{84.90} & \textbf{88.05} \\
 & 10  & 62.21 & 23.86 & 15.59 & 35.06 \\
 & 30 & 53.91 & 9.28 & 6.93 & 22.90 \\
\specialrule{0.4pt}{\aboverulesep}{\belowrulesep}
\multirow{3}{*}{GPT-o4-mini}
 & 1 & \textbf{92.34} & \textbf{92.45} & \textbf{93.44} & \textbf{92.74} \\
 & 10  & 88.11 & 76.61 & 73.94 & 79.55 \\
 & 30 & 75.44 & 76.31 & 14.86 & 55.54 \\
\bottomrule
\end{tabular}%
}
\caption{\textbf{RAG-style evaluation.} \textbf{ROUGE-1
    (R-1) recall} for our reasoning categories
  (world knowledge (W. Know.), arithmetic and temporal,
  averaged over Uni-speaker and Multi-speaker documents)
  and ``Average'' across categories.}
\label{tab:lc_metrics}
\end{table}

\paragraph{RAG Performance with an Oracle Retriever on Reason-Sensitive Documents.}
While retrieval quality clearly affects end-to-end performance, we ask whether an LLM with long-context capacity can still succeed \emph{once} the relevant document is present.
To test this, we use a retrieval-augmented generation (RAG) set-up with an \textbf{oracle retriever}, one that always includes the positive document in its top-\(k\).
The model sees the question together with \(k\) documents: one positive and \(k-1\) hard negatives sampled from the same pool (among other $M-1$ samples), ensuring comparable style and topic.
This configuration removes retrieval as a variable and isolates the LLM’s document-side reasoning ability.

We evaluate three settings: $k{=}1$ (positive only), $k{=}10$ (positive plus nine negatives), and $k{=}30$ (a full-pool setting where all documents from the pool are provided as context). The model receives the query along with the sequence of documents and must generate an answer.
We evaluate three reader models: \textsc{LLAMA~3.3 70B}, \textsc{GPT-4.1}\footnote{Checkpoint: \texttt{gpt-4.1-2025-04-14}}, and \textsc{GPT-o4-mini}\footnote{Checkpoint: \texttt{o4-mini-2025-04-16}}. In Table \ref{tab:lc_metrics}, we report the average ROUGE-1 recall\footnote{R-1 Rec. $=\frac{|\text{Output Unigrams } \cap \text{ Gold Answer Unigrams}|}{|\text{Gold Answer Unigrams}|}$} scores to measure the overlap between the generated output and the positive answer \cite{lin-2004-rouge}.
When given only the positive document ($k{=}1$), the models achieve  average ROUGE-1 Recall
of 81.92, 88.05, and 92.74.
This suggests that the query itself is straightforward to answer once the relevant document is isolated.
This also means that an LLM can solve the task if a high-performing retriever (which
would retrieve the relevant document at rank 1) is available.
However, as $k$ increases (even with the positive included), performance declines, showing that LLMs struggle to focus on the correct evidence amid structurally similar negatives. This supports prior findings on long-context limitations and highlights the need for retrieving a small, focused set of documents rather than increasing context size \cite{kuratov2024babilong, modarressi2025nolima}.

\paragraph{Error Analysis}
RAG has two stages---retrieval and generation---so we analyze errors along two axes.  
\textbf{1. Retrieval side (Rank-1 vs.\ Positive).} For each query, we compare the retriever’s top-1 passage with the annotated positive. We analyze the 60 queries where \textsc{Dragon+}’s top-1 document differs from the positive ($3$ reasoning categories $\times$ $2$ discourse styles $\times$ $10$ queries), yielding 120 passages (top-1 and positive per query). We categorize mis-rank reasons into four groups: (i) \emph{Word Overlap} (top-1 has extra query surface tokens), (ii) \emph{Semantic Cue} (similar overlap but extra topical/theme terms), (iii) \emph{Length} (overlap/semantics comparable; shorter passage chosen), and (iv) \emph{Unknown} (indistinguishable under our heuristics). Table~\ref{tab:retrieval_errors} shows that \emph{Semantic Cue} is most frequent in arithmetic queries, and \emph{Word Overlap} in temporal and world knowledge.  
\textbf{2. Generation side (Oracle-RAG, $k{=}10$ vs.\ all).} To isolate generation errors, we evaluate an oracle setting where the positive passage is guaranteed in context. We randomly select 60 queries ($3 \times 2 \times 10$) and evaluate two context sizes ($k{=}10$ and $k{=}\text{all}$, others selected randomly), yielding 120 cases. After reviewing outputs, we assign a single label to each incorrect answer: (i) \emph{Malformation} (positive present, answer malformed), (ii) \emph{No-answer/unrelated}, or (iii) \emph{Distraction} (copied from a surface-similar distractor). Table~\ref{tab:generation_errors} shows that \emph{No-answer/unrelated} is most frequent, \emph{Distraction} occurs mainly in temporal queries, and longer context reduces \emph{Malformation} but raises \emph{No-answer/unrelated}, suggesting that context length alone does not fix generation errors.

\begin{table}[t]
\centering
\small
\setlength{\tabcolsep}{6pt}
\resizebox{\columnwidth}{!}{%
\begin{tabular}{lcccc}
\toprule
\textbf{Reasoning} & \textbf{Word overlap} & \textbf{Semantic cue} & \textbf{Length} & \textbf{Unknown} \\
\midrule
Arithmetic      & 15\% & 55\% & 5\% & 25\% \\
Temporal        & 55\% & 5\%  & 35\% & 5\%  \\
W. Know. & 50\% & 5\%  & 30\% & 15\% \\
\bottomrule
\end{tabular}%
}
\caption{\textbf{Retrieval-side error types distribution for top-1 vs.\ positive.} For each reasoning category, we consider 20 query pairs (2 discourse styles $\times$ 10 queries); the percentages indicate the share of those 20 pairs for which the error type(column) was the primary reason the top-1 passage differed from the positive document (W. Know. = world knowledge).}
\label{tab:retrieval_errors}
\end{table}
\begin{table}[t]
\centering
\small
\setlength{\tabcolsep}{8pt}
\resizebox{\columnwidth}{!}{%
\begin{tabular}{llccc}
\toprule
\textbf{Reasoning} & \textbf{$k$} & \textbf{Malformation} & \textbf{No-answer/Unrelated} & \textbf{Distraction} \\
\midrule
Arithmetic       & 10   & 40\% & 60\% & 0\% \\
                 & all  & 20\% & 75\% & 5\% \\
\midrule
W. Know.  & 10   & 45\% & 55\% & 0\% \\
                 & all  & 15\% & 85\% & 0\% \\
\midrule
Temporal         & 10   & 0\%  & 35\% & 65\% \\
                 & all  & 0\%  & 40\% & 60\% \\
\bottomrule
\end{tabular}%
}
\caption{\textbf{Generation-side error type distribution under \emph{oracle} RAG with two context sizes ($k{=}10$ vs.\ $k{=}\text{all}$)}, where the positive document is included in the context. Results are based on a randomly selected set of 60 queries (120 evaluated cases). Percentages are computed over incorrect answers within each (reasoning category, $k$) cell (W. Know. = World Knowledge).}
\label{tab:generation_errors}
\end{table}

\section{Conclusion}

We introduce \impliret, a benchmark for evaluating retrieval
models when relevance depends on
document-side reasoning on
implicit
facts. Unlike prior datasets that emphasize
complex queries, \impliret shifts the reasoning burden to
the documents. It covers three reasoning types -- world
knowledge, arithmetic, and temporal -- and two discourse
styles. Across sparse, dense, and KG-augmented retrievers,
the best nDCG@10 is only 14.91. Even with GPT-o4-mini,
given thirty documents including the positive, performance peaks at
just
55.54\%. 
These results highlight the difficulty of
retrieving implicit facts and the need for models that can reason
beyond surface cues.

\section*{Limitations}

While our benchmark is carefully designed to evaluate implicit document-side reasoning in retrieval systems, it has the following limitations:

\paragraph{Synthetic Dataset.}
Documents and queries in \impliret are synthesized using LLMs and structured templates. This allows control over the facts and how they are implicitly expressed, while avoiding conflicts. It also enables easy regeneration if data contamination or memorization is suspected. As with any synthetic benchmark, the data may differ slightly from naturally occurring text in discourse structure or topic diversity. All examples are in English and follow conversational formats (uni-speaker chats and multi-speaker forum posts). Although the use of LLMs helps ensure fluency, it introduces the risk of subtle hallucinations or unintended cues, which we address through automatic verification during dataset construction.

\paragraph{Reasoning Types \& Level.}
In \impliret, we only cover three simple categories of reasoning relations: arithmetic, temporal, and world knowledge, each with shallow composition. While the coverage of reasoning types is limited, the core finding remains: current retrievers struggle to locate relevant documents when reasoning is implicit, and LLMs fail to reliably attend to the correct evidence in long-context settings.

\bibliography{acl_latex}

\appendix

\section{Dataset Generation}
\label{sec:appendix}

\noindent
In this appendix, we describe, for each of the three reasoning categories,
\emph{Arithmetic}, \emph{World-Knowledge}, and \emph{Temporal},
(i) how we construct the implicit-tuple sets and (ii) what arguments are
required to synthesize their corresponding contexts.
After covering tuple-set construction, we explain in
Section~\ref{sec:app_synthesizing_the_context} how a language model is
used to generate the final passages.

\subsection{Arithmetic Reasoning}
\label{sec:app_ds_gen_a}

For each implicit tuple set in the collection, to generate $M$ tuples for the Arithmetic category, we use Algorithm~\ref{alg:arith-uni}, which ensures that all tuples satisfy the these constraints: (i) all $p_{1}$ and $p_{2}$ values across the tuples are distinct, ensuring exactly one correct answer per query;
(ii) all $p_{1}$ and $p_{2}$ values should be a multiple of 10, so that values resemble realistic prices;
(iii) the \(p_{2}\) values  are evenly distributed across a predefined range of plausible prices, avoiding value clustering; and
(iv) the resulting multiplier must have at most 2 decimals; hence, the required calculation is simple.
The Algorithm\ref{alg:arith-uni} returns a set of tuples of the form \(((p_{1,i}, r_i, e_i), p_{2,i})\), where the two prices are mutually distinct and the multiplier \(r_i\) satisfies predefined numerical constraints. The construction guarantees uniform distribution across price ranges and ensures that each tuple encodes a plausible relative-price comparison suitable for reasoning-based retrieval.

\begin{algorithm}[t]
\caption{Arithmetic Tuple Set}
\label{alg:arith-uni}
\begin{algorithmic}[1]
\small
\Require Number of tuples \(M\); \textit{style} \(\in\{\text{multi},\text{uni}\}\); Total number of attempts \(limit\)
\If{\textit{style} $=$ multi}
    \State $(B_L,B_U)\gets(50,2050)$
\Else
    \State $(B_L,B_U)\gets(50,3050)$
\EndIf
\State $\Delta\gets\lfloor(B_U-B_L)/M\rfloor$
\State \text{ImplicitTupleSet} $\gets\varnothing$
\State \text{PriceSet} $\gets\varnothing$
\Repeat
    \For{$i\gets0$ \textbf{to} $M-1$}
        \State $p_{2,i}\gets B_L+i\Delta$
        \State $r_i\gets None$
        \State \textit{attempts}$\gets0$
        \Repeat
            \State Sample $p_{1,i}\sim\mathrm{Uniform}\bigl(\{x\in10\mathbb M\mid B_L\le $ \Statex\hspace{17.5em}  $ x\le B_U\}\bigr)$
            \If{$p_{2,i}$ > $p_{1,i}$}
                \State $r_i\gets\dfrac{p_{2,i}}{p_{1,i}}$ 
            \Else
                \State $r_i\gets\dfrac{p_{1,i}}{p_{2,i}}$ 
            \EndIf
            \State \textit{attempts}$\gets$\textit{attempts}$+1$
        \Until{ ($0<r_i<3$ \textbf{and} Round($r_i$,2) = $r_i$ \textbf{and} 
        \Statex\hspace{5.5em} $ p_{1,i}\notin PriceSet$ and $p_{2,i}\notin PriceSet$)
        \Statex\hspace{5.5em} \textbf{or} \textit{attempts} exceeds \textit{limit}}
        \If{$r_i$}
            \State $e_i\gets\begin{cases}\texttt{Lower}, & p_{2,i}<p_{1,i}\\
                                           \texttt{Higher},&\text{otherwise}\end{cases}$
            \State $\text{ImplicitTupleSet}\gets\text{ImplicitTupleSet}$ 
            \Statex\hspace{12em} $ \cup\{((p_{1,i},r_i,e_i),p_{2,i})\}$
        \Else
            \State $p_{2,i}\gets p_{2,i}+1$
            \If{$p_{2,i}= B_L + (i+1)\Delta$}
                \State \textbf{restart entire generation}
            \EndIf
        \EndIf
    \EndFor
\Until{|\text{ImplicitTupleSet}| $=M$}
\State \Return \text{ImplicitTupleSet}
\end{algorithmic}
\end{algorithm}

\subsection{World-Knowledge Reasoning}
\label{sec:ds_gen_s}
As described in Section~\ref{sec:dataset_generation}, we gather landmark-country pairs under three constraints:
(i) each landmark must refer to a globally unique location, avoiding names that could correspond to multiple places;
(ii) the landmark name must not include lexical, semantic, or language-specific cues that reveal its country, avoiding surface-form shortcuts; and
(iii) landmarks must refer to specific, recognizable sites rather than generic ones. To do so, we first assemble a seed list of unique landmark-country pairs via five steps: (i) issue a SPARQL query to Wikidata to retrieve every entity whose \texttt{instance-of} (P31) chain includes exactly one of the high-level place classes, museum (\texttt{Q33506}), university (\texttt{Q3918}), church building (\texttt{Q16970}), venue (\texttt{Q17350442}), or landmark (the superclass of \texttt{Q17350442}), and that is linked to exactly one sovereign state via the \texttt{country} (P17) property;  (ii) for each hit, extract the English labels of its enclosing administrative region (P131), city (P131 restricted to \texttt{Q515}), generic location (P276), and street (P669), yielding up to five concentric location strings;  (iii) discard entities with missing, machine-generated, or multi-country labels, then drop any whose name tokens overlap these location strings; embed each remaining landmark-country pair with the 768-dimensional Contriever encoder and retain only those with cosine similarity below 0.25; (iv) pass the remainders to a 70B-parameter Llama-3.3 classifier that flags and removes generic names (e.g., “Downtown Club”) or labels leaking their country, using exponential back-off retries until accepted; and (v) submit each accepted label to GPT-4o in web-search mode\footnote{Checkpoint: \texttt{gpt-4o-search-preview-2025-03-11}}, prompting it to return the place’s country in a dictionary format, and keep only those for which the model returns exactly one location.  This process yields a balanced pool of 100 unique landmark-country pairs, ensuring we sample across different countries rather than multiple landmarks from the same country. 
Finally, for each implicit tuple set, we select a set of $M$ distinct countries \(C\) and, for each \(c\in C\), sample one landmark \(l_c\sim\mathrm{Uniform}(L_c)\) from that country’s filtered list \(L_c\). 
The resulting \emph{implicit tuple sets} is as follows:
$$
\textit{ImplicitTupleSets} \;=\; \{\, (l_c,\,c) \mid c\in C \,\}.
$$
Generating $N$ implicit tuple sets, we have our collection.

\subsection{Temporal Reasoning}
\label{sec:ds_gen_t}

As described in Section~\ref{sec:dataset_generation}, for each tuple set, we generate a implicit tuple sets of \(M\) tuples \(\bigl((d_{B},R),D_{L}\bigr)\) to have the following constraints: (i) all explicit dates in any $D_L$ are unique within the tuple set, ensuring that each query maps to exactly one positive document;
(ii) all dates used as base or resolved targets are evenly distributed across a fixed date window to avoid clustering; and 
(iii) all relative offsets in $R$ must fall within a limited number of days from. Because context synthesis differs between multi-speaker and uni-speaker modes, we describe the two procedures separately in the following subsections.

\subsubsection{Multi-Speaker: Tuple Construction}
\label{sec:multi_temporal_algo}

Here, we generate $N$ implicit tuple lists, each containing $M$ tuples. For generating them, we use Algorithm~\ref{alg:multi-temporal}. The returned output contains all the information detailed before.

\begin{algorithm}[t]
\caption{Multi-Speaker Temporal Tuple Set}
\label{alg:multi-temporal}
\begin{algorithmic}[1]
\small
\Require $DateWindow$ from 2024-01-01 to 2024-12-31
\Statex\hspace{1em}\textproc{DateSelection}$(n, DateWindow)$: return $n$ distance \textit{date} in $DateWindow$ in which for $i \in [0,\dots,n-2]$, days between $date_i$ and $date_{i+1}$ be between 2, up to 7 days, and the distance between $date_{n-1}$ and the end of $DateWindow$ be at least 14 days.
\Statex
\State $\{\text{work\_date}_i\}_{i=0}^{19}\gets$ 
\Statex\hspace{6em} \textproc{DateSelection}$(20,DateWindow)$
\For{$i\gets0$ \textbf{to} $18$}                      
    \State $\text{message\_date}_i\gets$ $\text{work\_date}_{i+1}$
    \State $r_i\gets(\text{message\_date}_i-\text{work\_date}_i).\text{days}$
    \State $TupleSet\gets TupleSet$ 
    \Statex\hspace{7em} $\cup((\text{message\_date}_i,r_i),\text{work\_date}_i)$
\EndFor
\State $\text{message\_date}_{19}\gets$ $\text{work\_date}_{19}$ + 
\Statex\hspace{11em} \textproc{UniformSample}$((1,7))$
\State $r_{19}\gets(\text{message\_date}_{19}-\text{work\_date}_{19}).\text{days}$
\State $TupleSet\gets TupleSet$ 
\Statex\hspace{5em} $\cup((\text{message\_date}_{19},r_{19}),\text{work\_date}_{19})$
\State \Return $TupleSet$
\end{algorithmic}
\end{algorithm}

\subsubsection{Uni-Speaker: Tuple Construction}
\label{sec:uni_temporal_algo}

We want to generate $N$ implicit tuple sets, each containing $M$ tuples. Each tuple set describes a main conversant’s 28-day activity schedule.  
We categorize activities into three types:
\begin{enumerate}
  \item \textbf{One-time:} executed exactly once in the 14-day period (we have 9 for then in each schedule).
  \item \textbf{Repeating-Non-Sequential:} occurring on multiple, \emph{non-consecutive} days (we have 3 for them in each schedule: 2 days, 3 days, and 2 days).
  \item \textbf{Repeating-Sequential:} performed on consecutive days (we have 3 for them in each schedule: 3 days, 3 days, and 4 days).
\end{enumerate}
Each set covers two consecutive 14-day blocks, contains exactly \(M \)=30 activities, and is constructed in three phases:  
(i) scheduling activities without temporal overlap,  
(ii) selecting one \emph{message time} per activity such that it differs from the scheduled lot, and  
(iii) packaging every activity into a tuple  
\((\text{day(s)},\text{start\_hour},\text{end\_hour}, \text{message\_time})\).  
Algorithm~\ref{alg:uni-temporal} details the procedure.

\begin{algorithm}[t]
\caption{Uni–Speaker Temporal Tuple Set}
\label{alg:uni-temporal}
\begin{algorithmic}[1]
\small
\Require period duration \(14\)-days, day span \(7{:}00\text{--}19{:}00\)
\Statex\textbf{Auxiliary functions}

\Statex\hspace{1em}\textproc{PlaceSchedule}$(d,\;T,\;F,\;S)$: place a 2-4 hour scheduled slot, if $T=\text{`Seq'}$, for \emph{d} consecutive days, if $T=\text{`NonSeq'}$, for \emph{d} non-consecutive days, and if $T=\text{`Once'}$, for one day; mark blocks in $F$ to remove the free times, add the slot into the schedule list $S$, and make a tuple from the list of occupied days \(D_L\) and start and end hours (\((h_{start}, h_{end})\)) as \(a=(D_{La}, h_{start},h_{end})\) and returns it
\Statex\hspace{1em}\textproc{ShiftDates}(S,$d_{\mathrm{off}}$): shift all the relative days to 14 days later if $d_{\mathrm{off}}$=1.
\Statex \hspace{1em}\textproc{SelectQTimes}(S): Randomly selects a random day and hour (not exact time of start or end) as question time for each schedule to appear in the query. 

\Statex \hspace{1em}\textproc{DiffTimes}(m, a): Returns a list of day differences between every scheduled day \(d_{i,a}\) in the activity \(a\) and \(m\) (\((m- d_{i,a}).day\)).

\State $S\gets\varnothing$ \Comment{Schedule list}
\State $A\gets\varnothing$ \Comment{Activity list}
\For{$period\in\{0,1\}$} 
    \Statex \Comment{two consecutive 14-day blocks}
    \State $F\gets\{d\mapsto$\Call{InitFree}{$7,19$}$\mid d=1{:}14\}$ 
    \Statex \Comment{free-time map}
    \State $d_{\mathrm{off}}\gets 14\cdot period$     \Comment{calendar shift}
    \ForAll{$\mathrm{len}\in\{3,3,4\}$}
        \State \(A \gets A \, \cup\)\Call{PlaceSchedule}{$\mathrm{len},\textit{`Seq'},F,S$}
    \EndFor
    \ForAll{$\mathrm{len}\in\{2,2,3\}$}
        \State \(A \gets A \, \cup\)\Call{PlaceSchedule}{$\mathrm{len},\textit{`NonSeq'},F,S$}
    \EndFor
    \For{$i\gets1$ \textbf{to} $9$}
        \State \(A \gets  A \, \cup\)\Call{PlaceSchedule}{$\mathrm{len},\textit{`Once'},F,S$}
    \EndFor
    \State \Call{ShiftDates}{S,$d_{\mathrm{off}}$}
\EndFor
\State $\mathcal{Q}\gets$ \Call{SelectQTimes}{S}
\State \textit{TupleSet}$\gets\varnothing$
\ForAll{activity $a\in A$}
    \State $m_a\gets$ \Call{Random}{$\mathcal{Q}\setminus\{\text{times}(a)\}$}
    \State $R_a \gets $ \Call{DiffTimes}{$m_a,a$}
    \Statex \Comment{Message time}
    \State  \( TupleSet \gets TupleSet\) 
    \Statex\hspace{5.5em} \( \cup ( (m_a, R_a), a) \)
\EndFor
\State \Return \textit{TupleSet}
\end{algorithmic}
\end{algorithm}

\subsection{Synthesizing the Context}
\label{sec:app_synthesizing_the_context}
Now, for each reasoning category and document style, we have a list of implicit tuple sets containing all the information needed to have a consistent dataset. Consider having the auxiliary lexical content (personal names, daily-work verbs, brand names with corresponding items, and per-category forum topics and questions needed)\footnote{Generated using Gemma-3-27B-it \cite{gemma_2025}}, to each tuple of the implicit tuple list, we assign unique entities and then each tuple, contains all the information to generating the document in natural way. The exact item required for each reasoning category and style is mentioned in Table~\ref{tab:tracks-styles-assign}. Depending on the style, we proceed as follows:

\subsubsection{Uni-Speaker (Chat-Style)}
\begin{figure*}[t]
    \centering
    \includegraphics[width=1.0\textwidth]{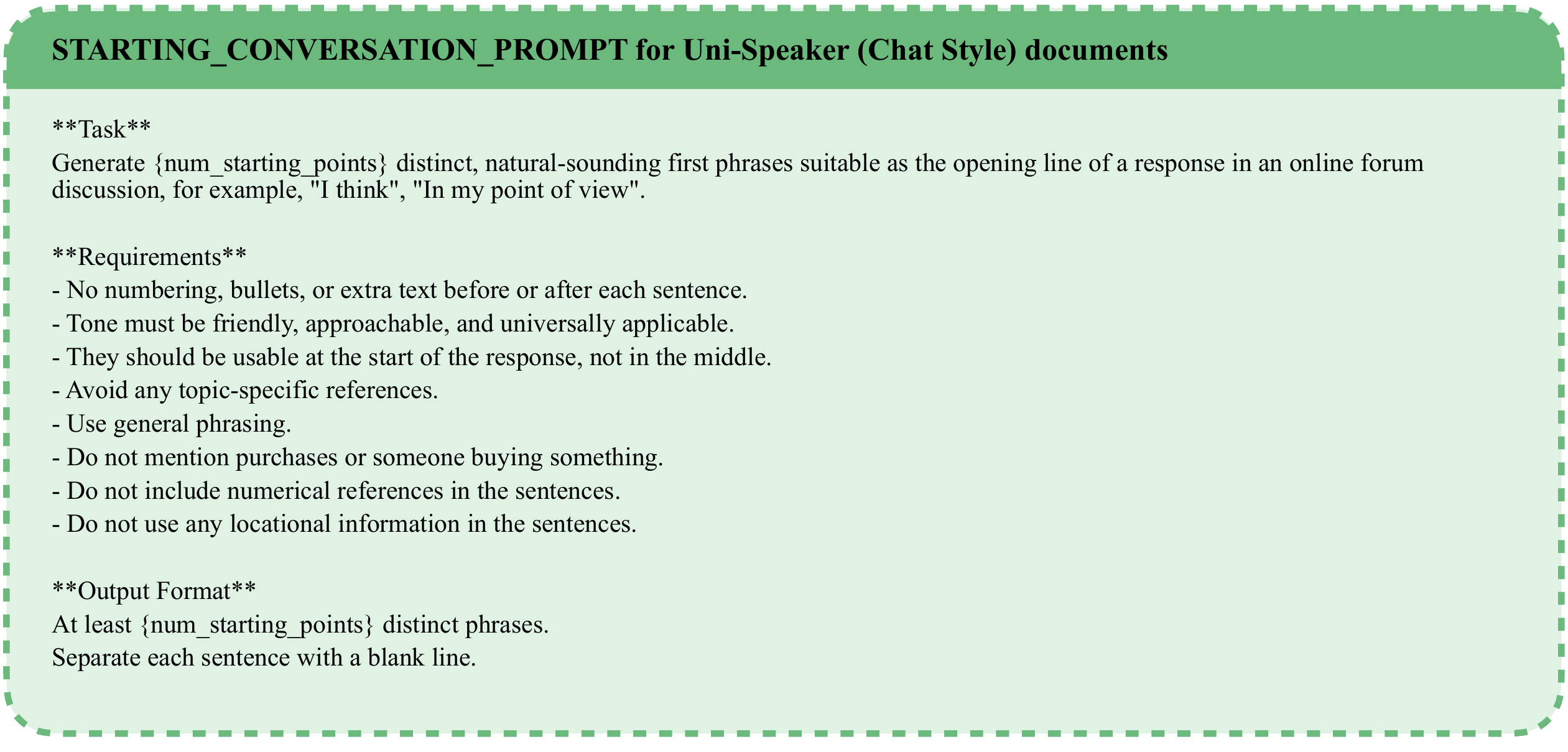}
    \caption{Prompt for generating a list of the first phrase in Uni-Speaker (Chat Style) documents. This prompt is used for all the reasoning categories of the Arithmetic, World-Knowledge, and Temporal.}
    \label{fig:11_Starting_conv_uni}
\end{figure*}

\begin{figure*}[t]
    \centering
    \includegraphics[width=1.0\textwidth]{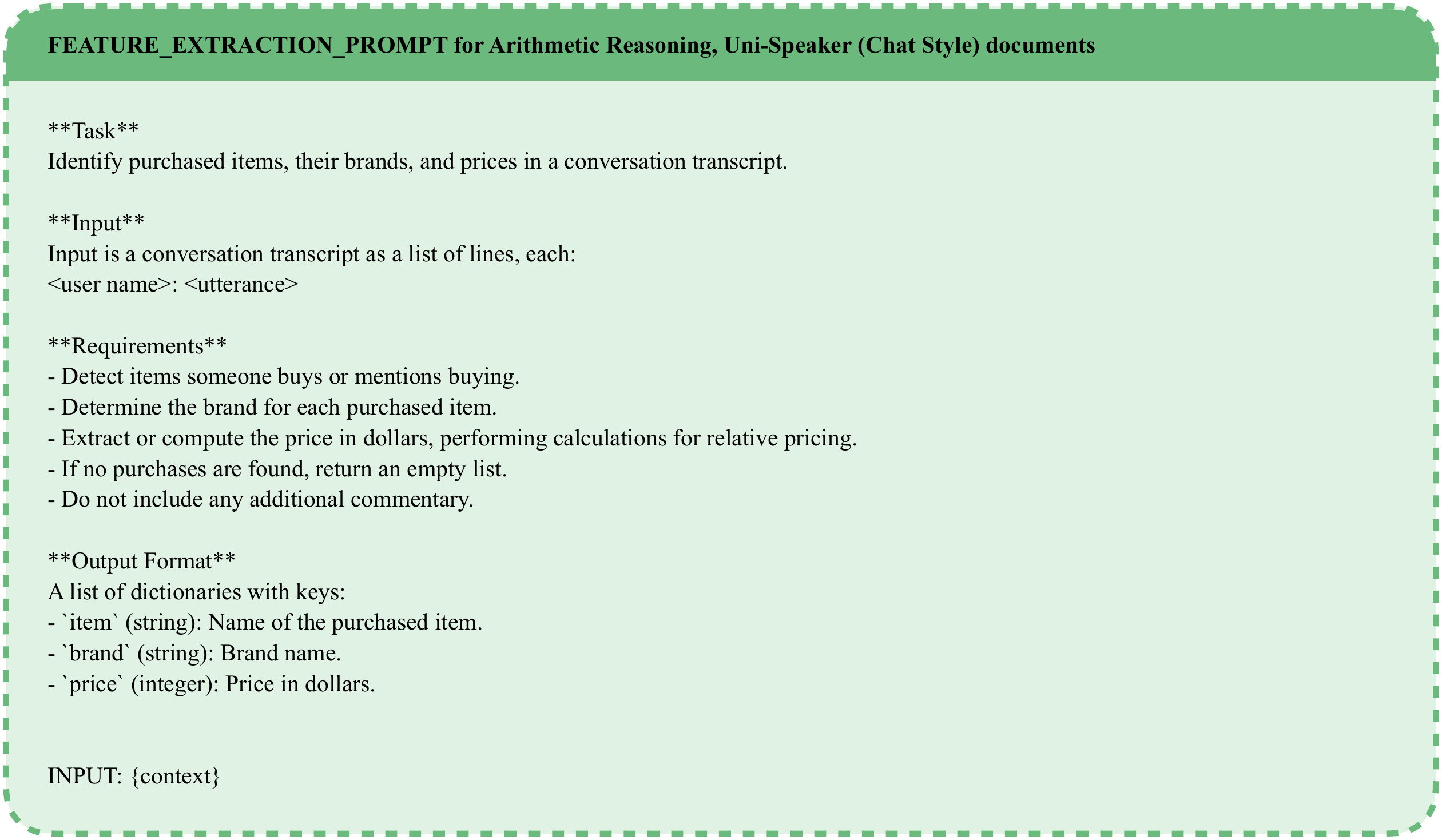}
    \caption{Prompt for reconstructing the original tuple of implicit tuple set (extracting features) from generated conversations for Multi-Speaker (Forum Style) documents in Arithmetic reasoning.}
    \label{fig:12_feature_extract_arith_uni}
\end{figure*}

\begin{figure*}[t]
    \centering
    \includegraphics[width=1.0\textwidth]{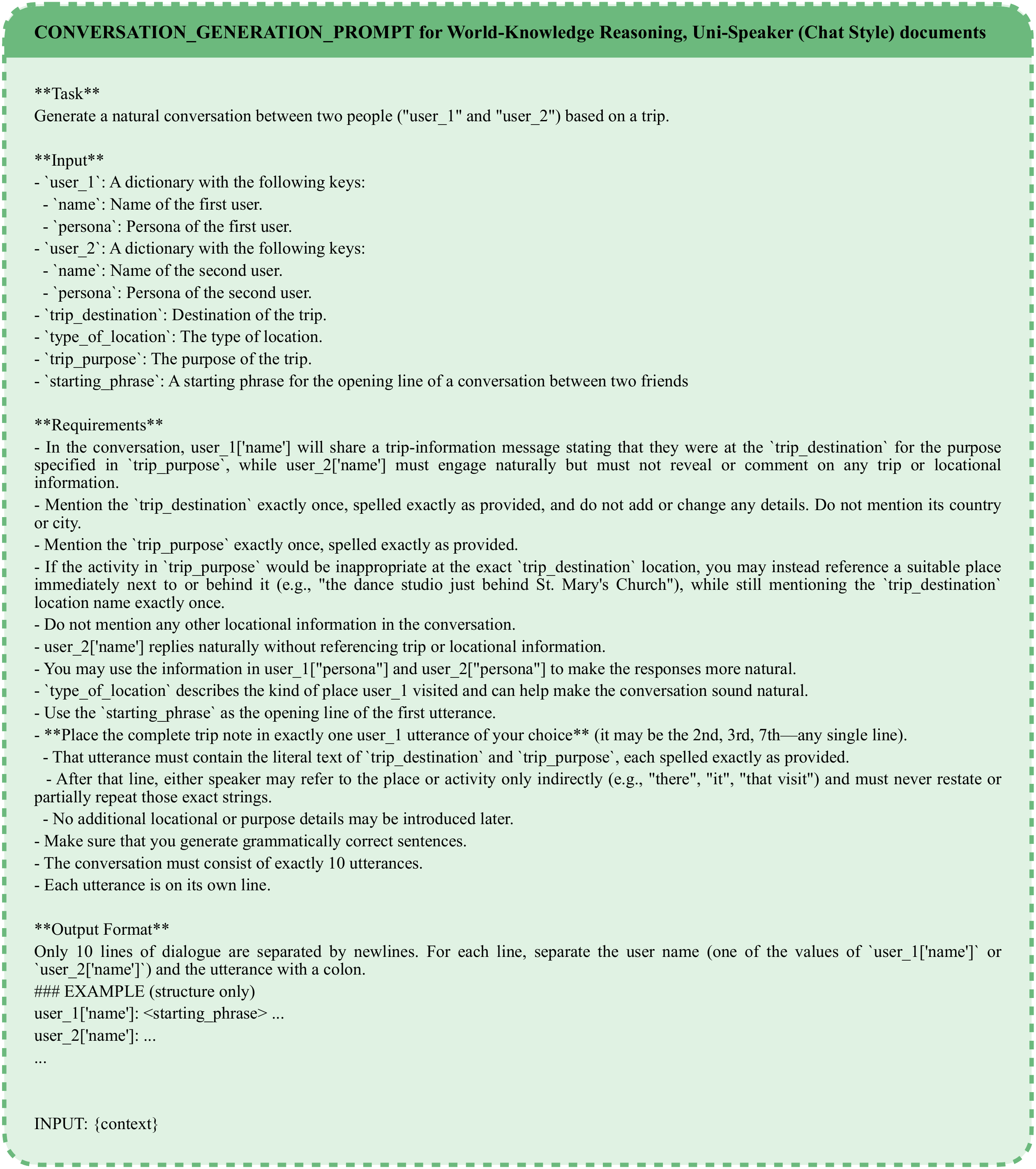}
    \caption{Prompt for generating the conversations for Uni-Speaker (Chat Style) documents in the World-Knowledge reasoning.}
    \label{fig:13_conv_gen_semantic_uni}
\end{figure*}

\begin{figure*}[t]
    \centering
    \includegraphics[width=1.0\textwidth]{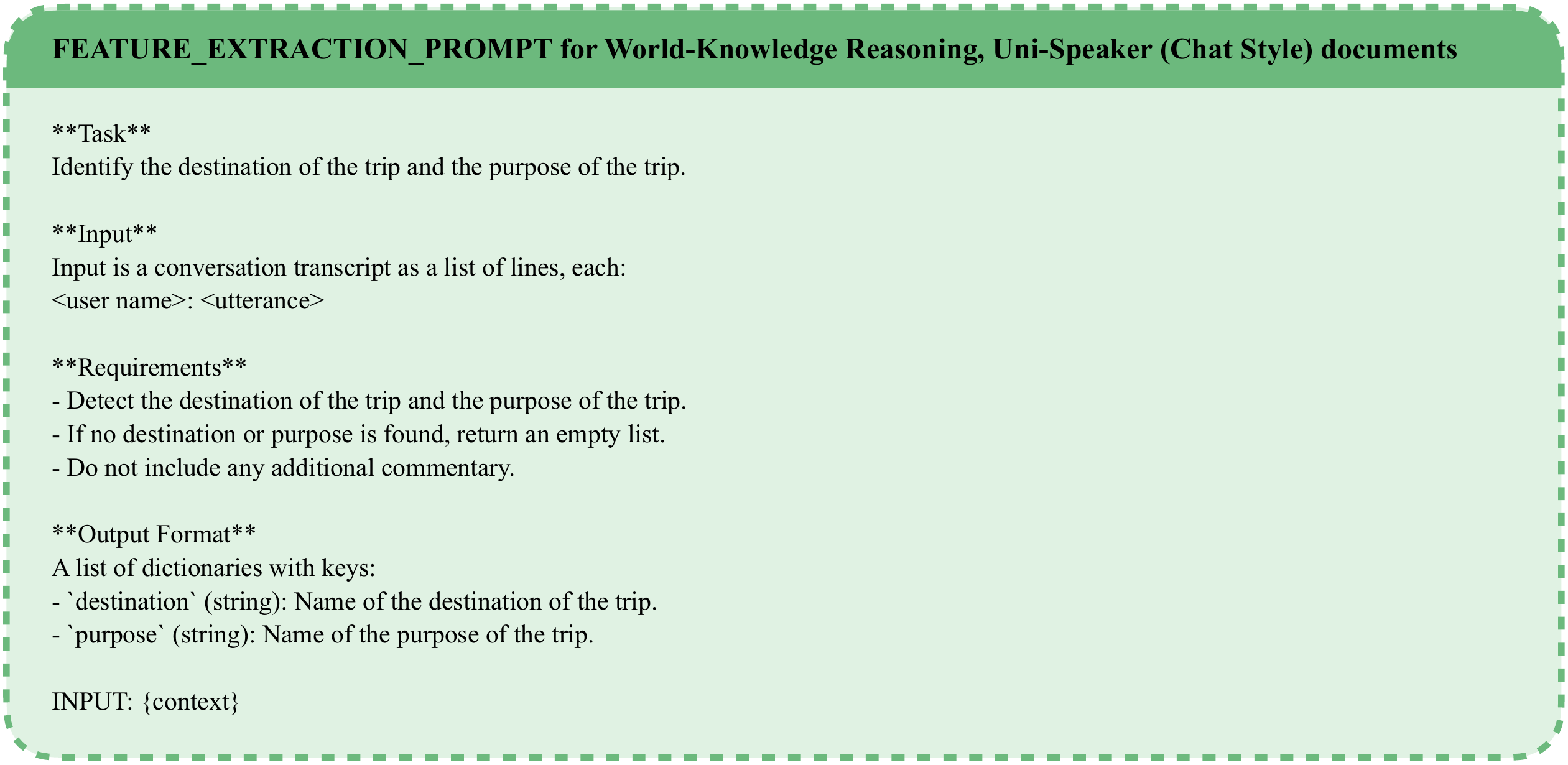}
    \caption{Prompt for reconstructing the original tuple of implicit tuple set (extracting features) from generated conversations for Uni-Speaker (Chat Style) documents in the World-Knowledge reasoning.}
    \label{fig:13_feature_extract_semantic_uni}
\end{figure*}

\begin{figure*}[t]
\centering
\includegraphics[width=1.0\textwidth]{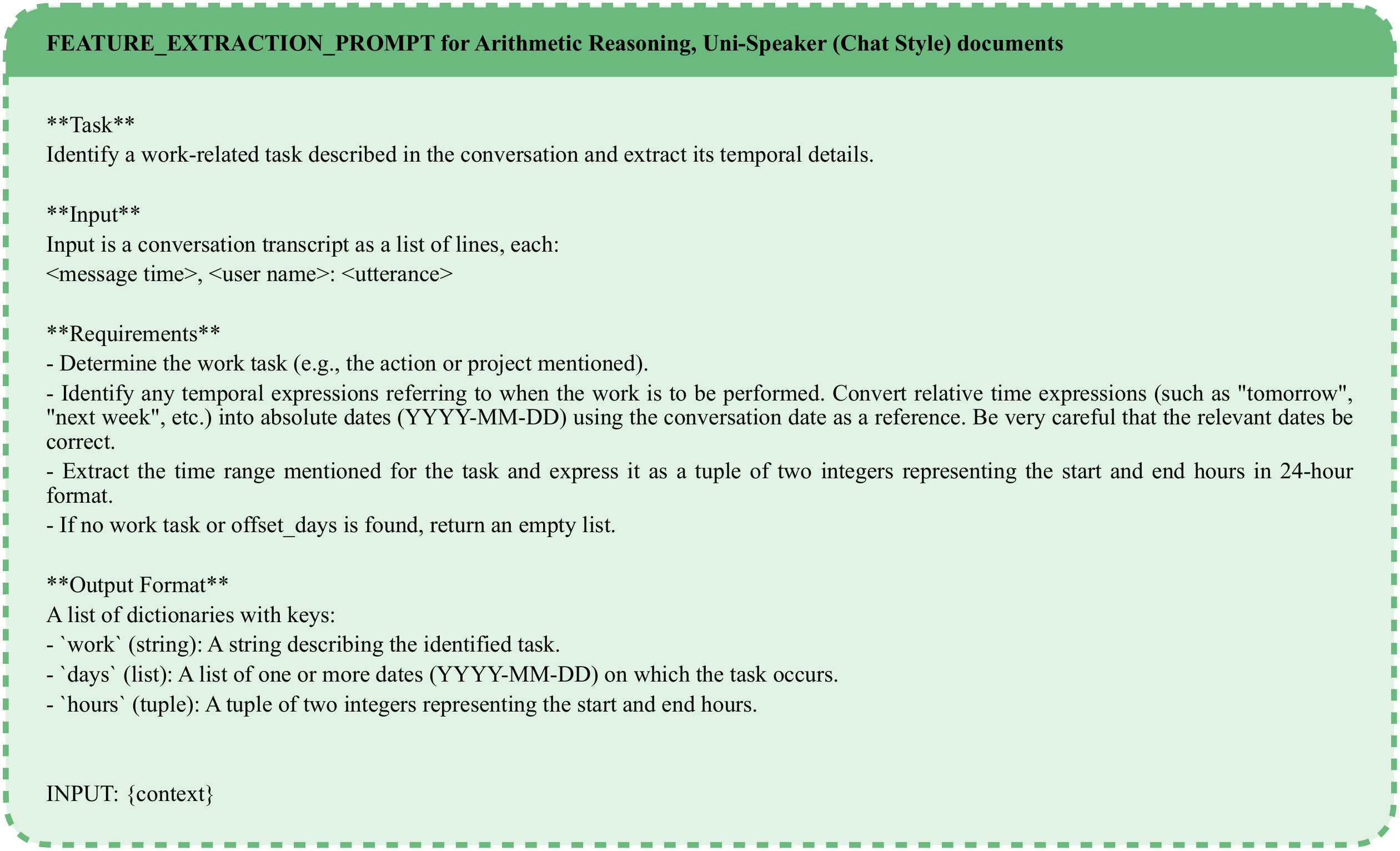}
\caption{Prompt for reconstructing the original tuple of implicit tuple set (extracting features) from generated conversations for Uni-Speaker (Chat Style) documents in the Temporal reasoning.}
\label{fig:14_feature_extract_temporal_uni}
\end{figure*}

In the conversation-generation stage, we load the implicit tuple sets for each (category, style) pair.  First, we use the \texttt{STARTING\_CONVERSATION\_PROMPT} to generate $M=30$ unique “starting phrases” for the tuple set (one for each tuple), ensuring that no two dialogues begin identically (Figure~\ref{fig:11_Starting_conv_uni}). Next, for each tuple, we prepend one of these starting phrases and feed the combination into the \texttt{CONVERSATION\_GENERATION\_PROMPT}; category-specific prompt templates and requirements are shown in Figure~\ref{fig:12_conv_gen_arith_uni} for the Arithmetic, Figure~\ref{fig:13_conv_gen_semantic_uni} for the World-Knowledge, and Figure~\ref{fig:14_conv_gen_temporal_uni} for the Temporal reasoning.  We then ask the model to produce exactly ten utterances per chat, verify the count, and regenerate any that do not meet this criterion. Finally, each completed conversation is submitted to a separate LLM via the \texttt{FEATURE\_EXTRACTION\_PROMPT}, which must reconstruct the original tuple to confirm that the dialogue faithfully conveys the intended information (Figure~\ref{fig:12_feature_extract_arith_uni} for the Arithmetic, Figure~\ref{fig:13_feature_extract_semantic_uni} for the World-Knowledge, and Figure~\ref{fig:14_feature_extract_temporal_uni} for the Temporal reasoning).

\subsubsection{Multi-Speaker (Forum-Style)}
\begin{figure*}[t]
    \centering
    \includegraphics[width=1.0\textwidth]{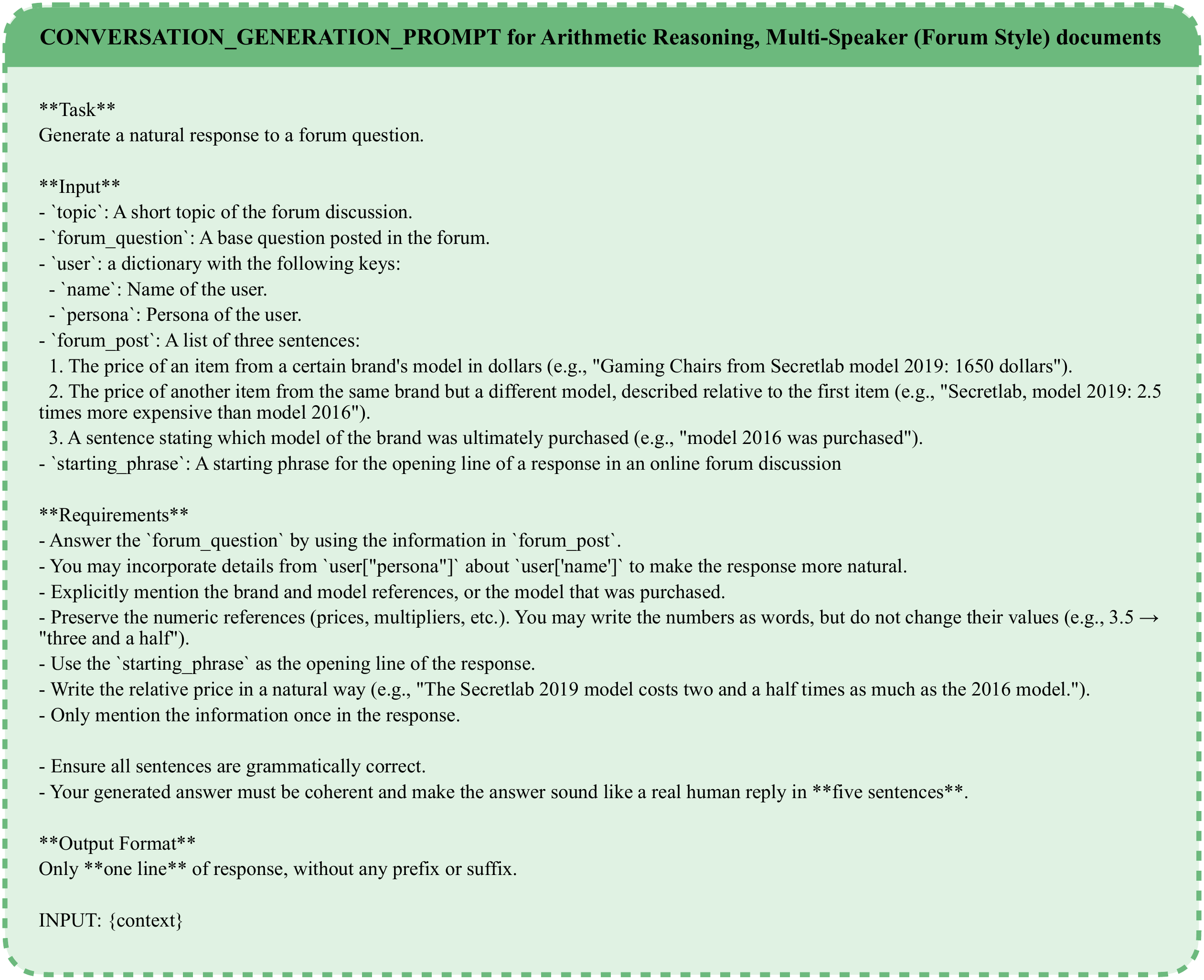}
    \caption{Prompt for generating the conversations for Multi-Speaker (Forum Style) documents in the Arithmetic reasoning.}
    \label{fig:12_conv_gen_arith_multi}
\end{figure*}

\begin{figure*}[t]
    \centering
    \includegraphics[width=1.0\textwidth]{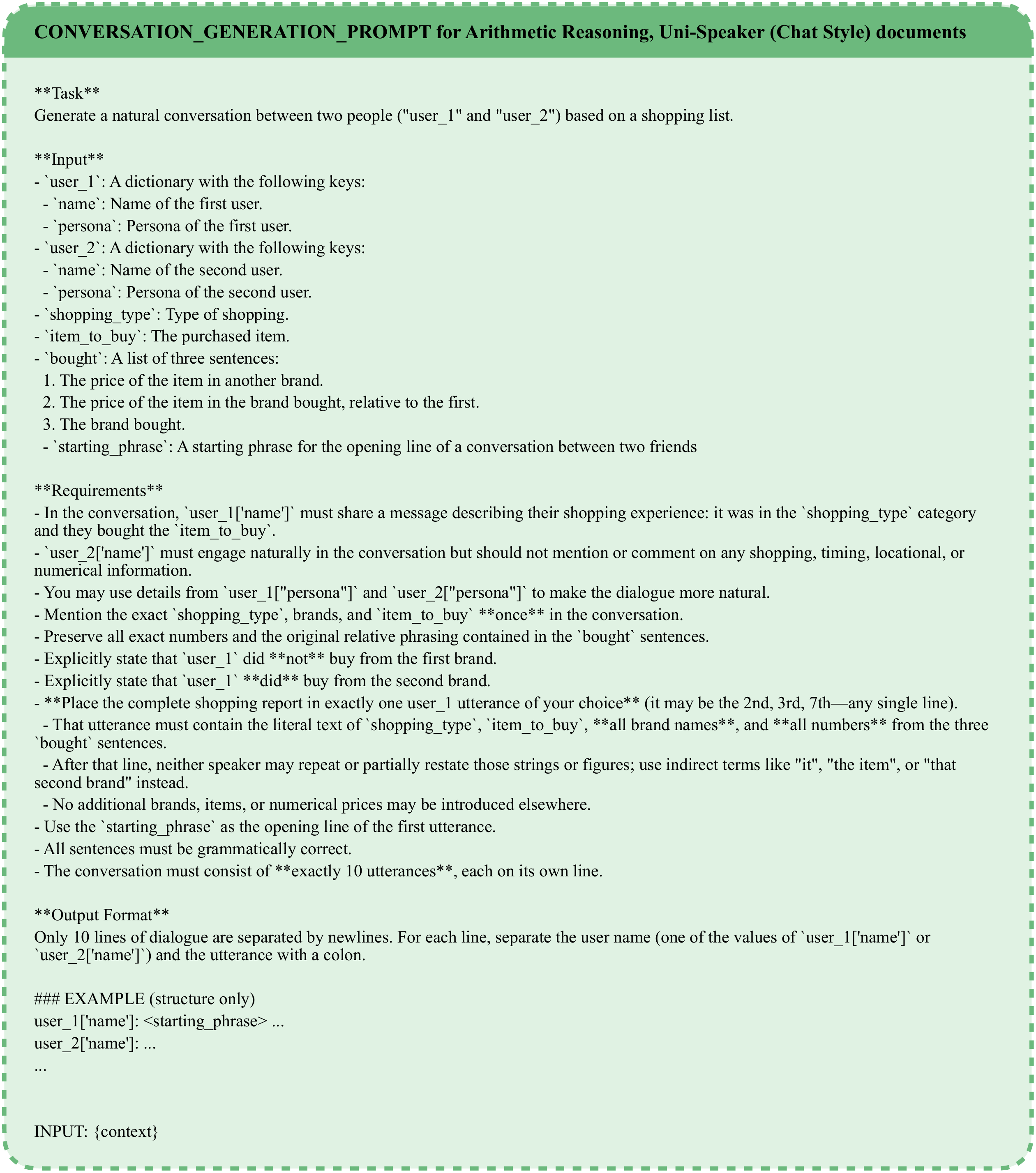}
    \caption{Prompt for generating the conversations for Uni-Speaker (Chat Style) documents in the Arithmetic reasoning.}
    \label{fig:12_conv_gen_arith_uni}
\end{figure*}

\begin{figure*}[t]
    \centering
    \includegraphics[width=1.0\textwidth]{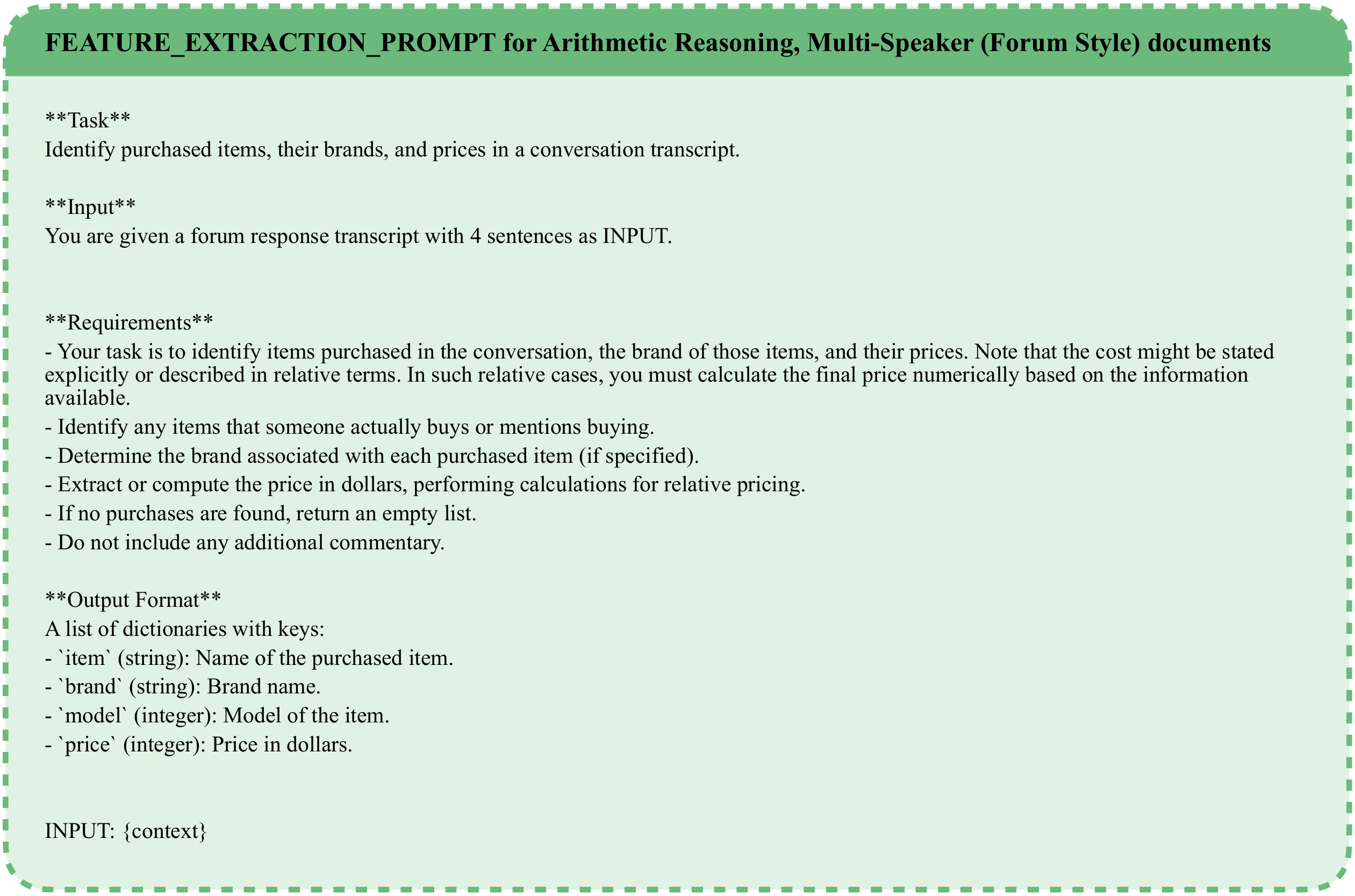}
    \caption{Prompt for reconstructing the original tuple of implicit tuple set (extracting features) from generated conversations for Multi-Speaker (Forum Style) documents in Arithmetic reasoning.}
    \label{fig:12_feature_extract_arith_multi}
\end{figure*}

\begin{figure*}[t]
    \centering
    \includegraphics[width=1.0\textwidth]{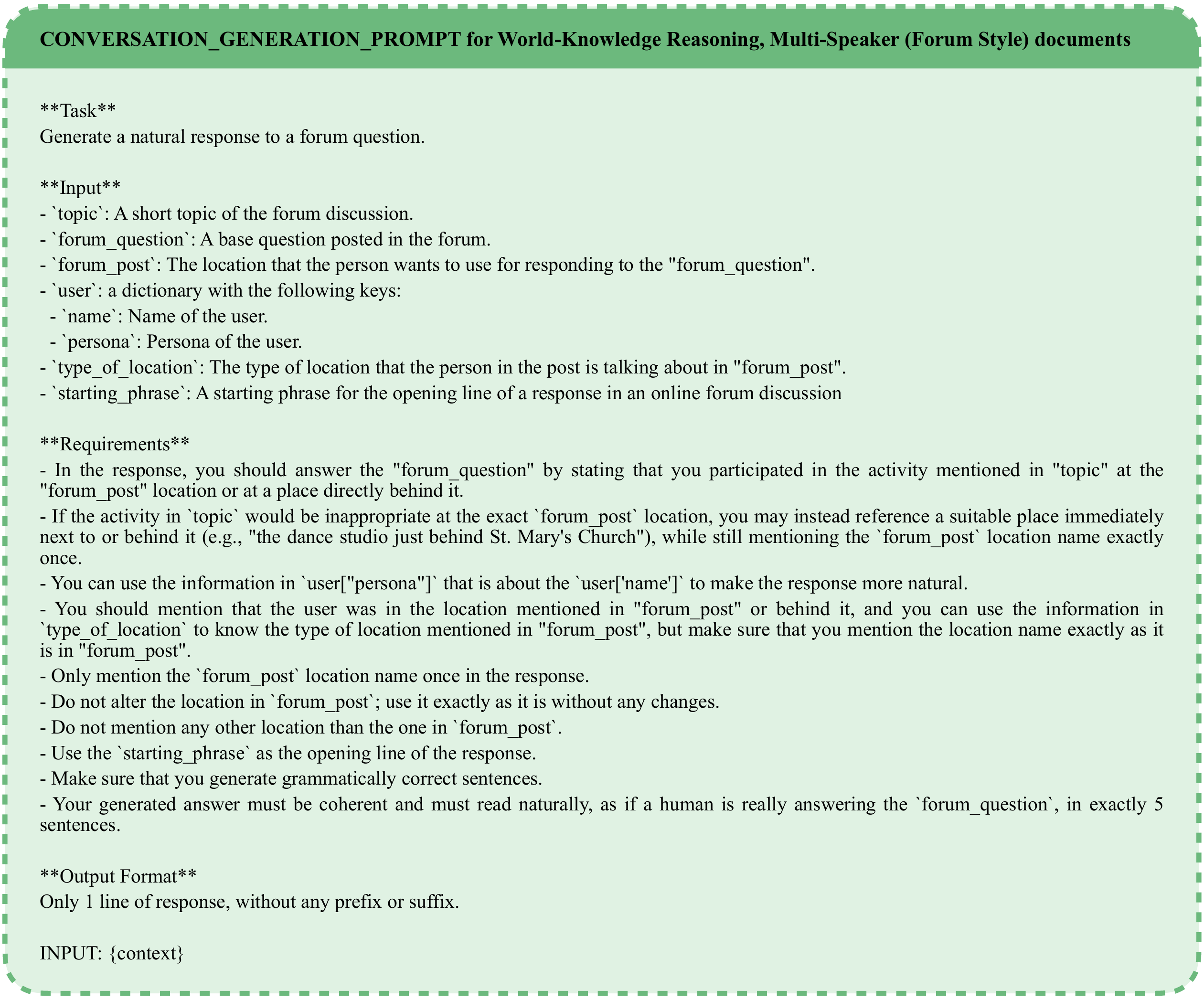}
    \caption{Prompt for generating the conversations for Multi-Speaker (Forum Style) documents in the World-Knowledge reasoning.}
    \label{fig:13_conv_gen_semantic_multi}
\end{figure*}

\begin{figure*}[t]
    \centering
    \includegraphics[width=1.0\textwidth]{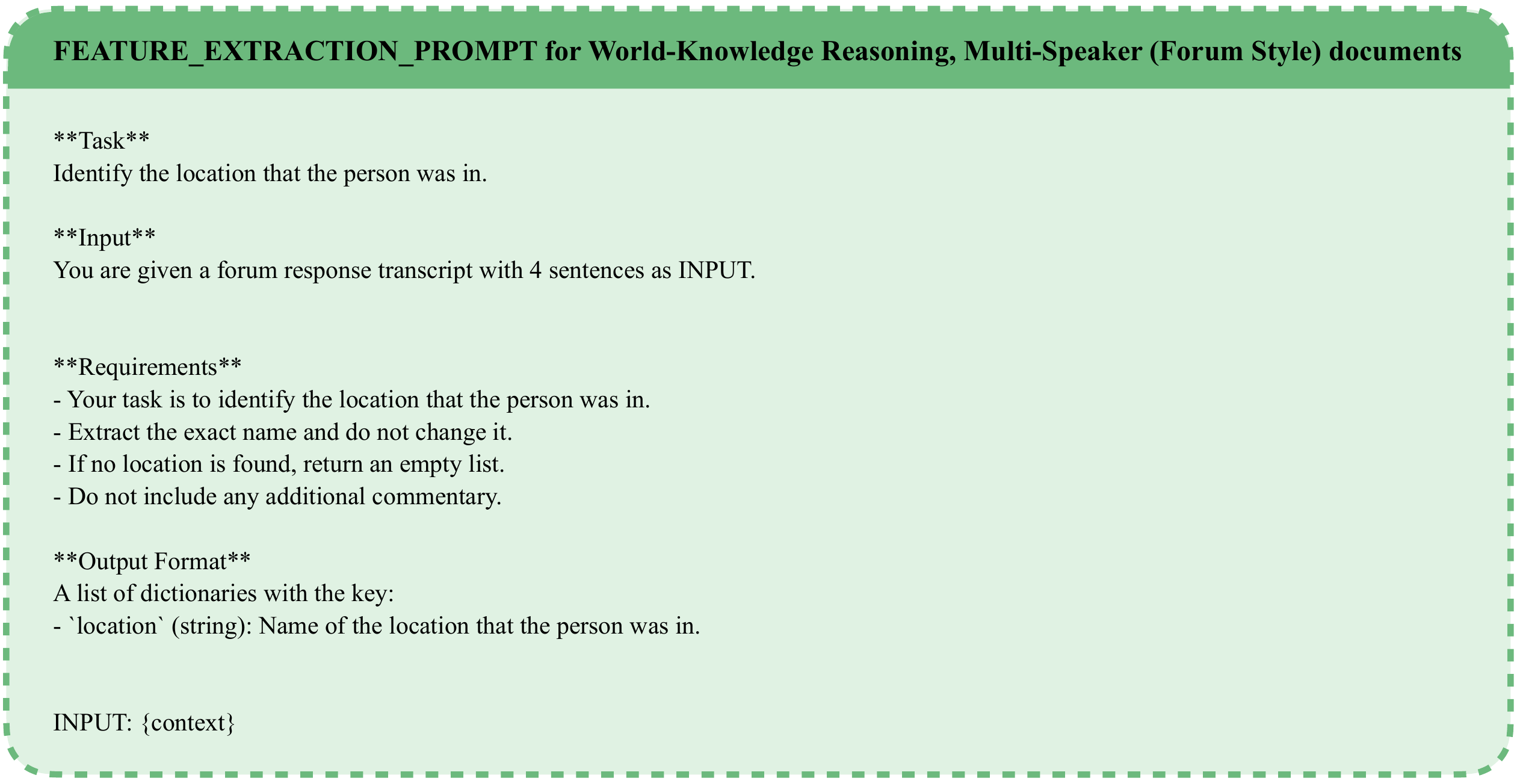}
    \caption{Prompt for reconstructing the original tuple of implicit tuple set (extracting features) from generated conversations for Multi-Speaker (Forum Style) documents in the World-Knowledge reasoning.}
    \label{fig:13_feature_extract_semantic_multi}
\end{figure*}

\begin{figure*}[t]
    \centering
    \includegraphics[width=1.0\textwidth]{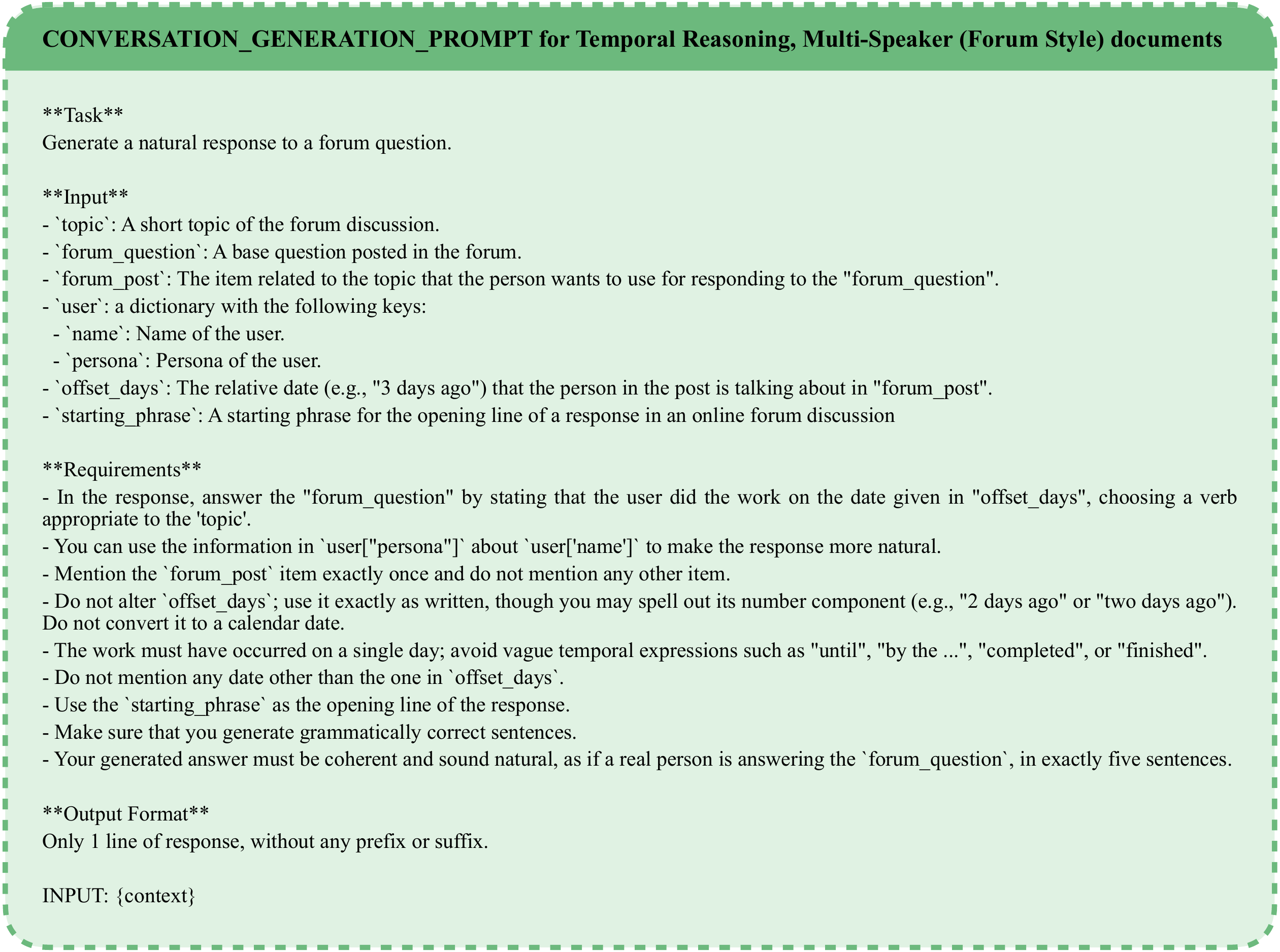}
    \caption{Prompt for generating the conversations for Multi-Speaker (Forum Style) documents in the Temporal reasoning.}
    \label{fig:14_conv_gen_temporal_multi}
\end{figure*}

\begin{figure*}[t]
    \centering
    \includegraphics[width=0.9\textwidth]{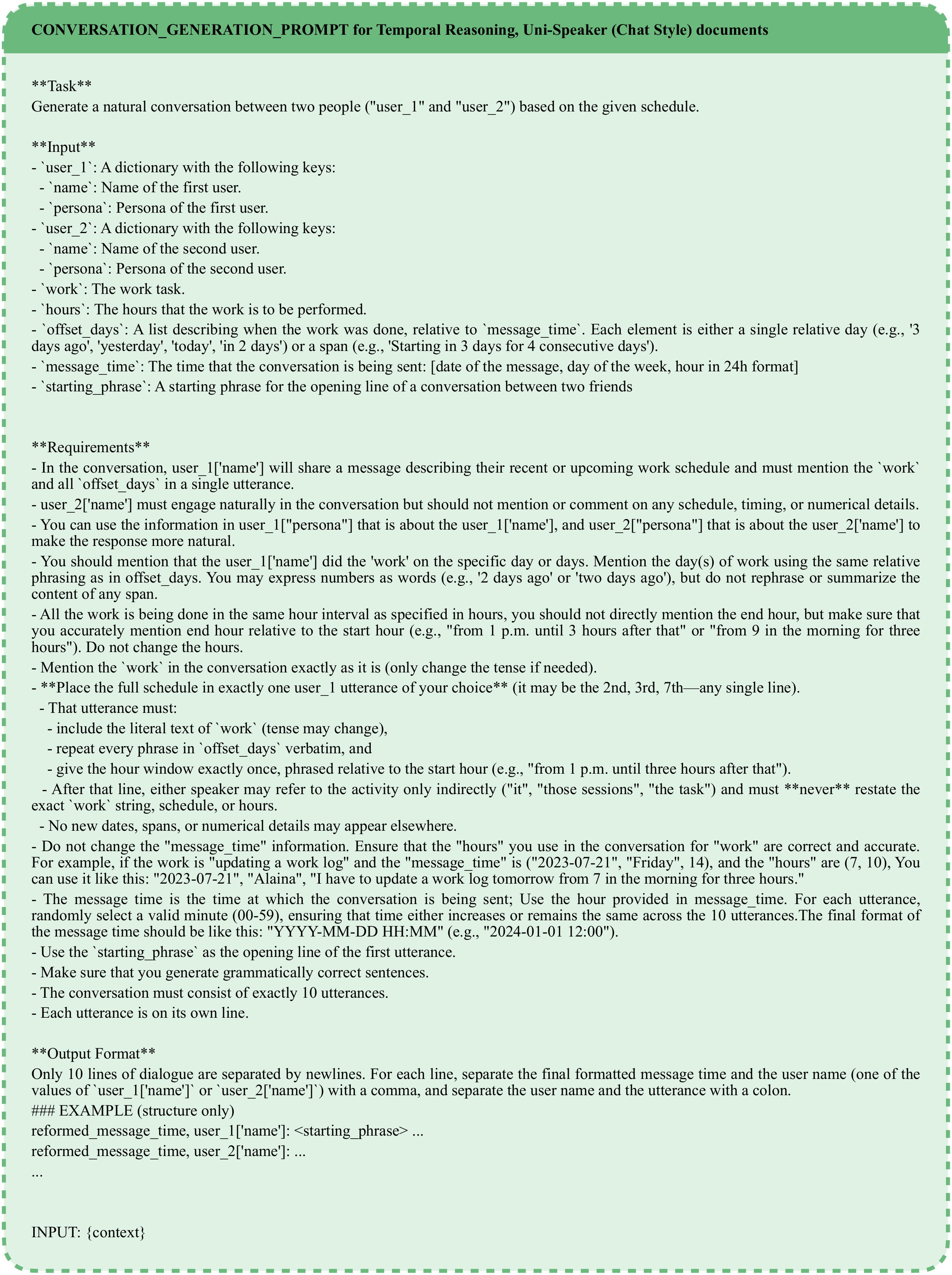}
    \caption{Prompt for generating the conversations for Uni-Speaker (Chat Style) documents in the Temporal reasoning.}
    \label{fig:14_conv_gen_temporal_uni}
\end{figure*}

\begin{figure*}[t]
    \centering
    \includegraphics[width=1.0\textwidth]{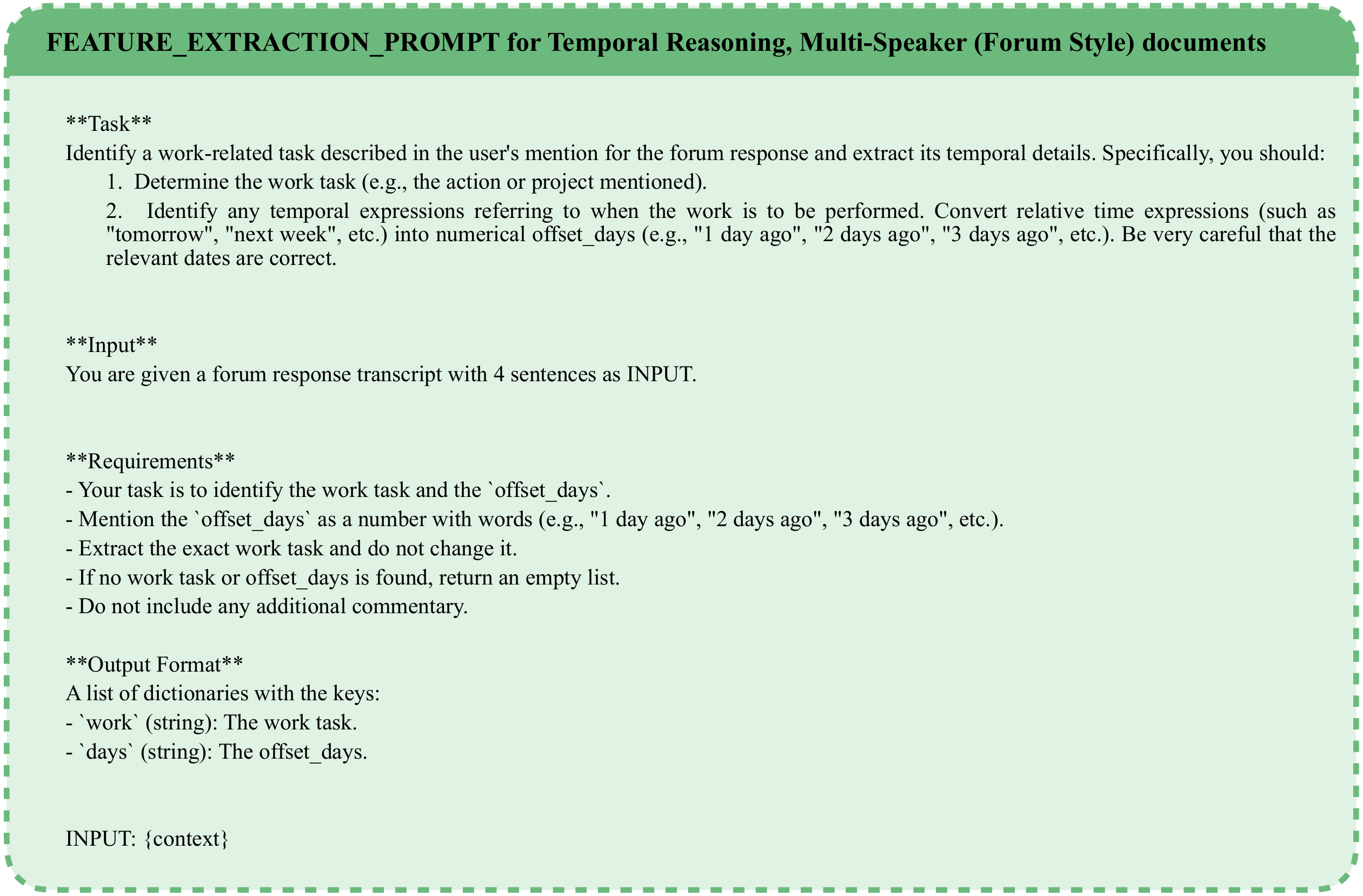}
    \caption{Prompt for reconstructing the original tuple of implicit tuple set (extracting features) from generated conversations for Multi-Speaker (Forum Style) documents in the Temporal reasoning.}
    \label{fig:14_feature_extract_temporal_multi}
\end{figure*}

In the forum-style generation stage, we similarly load the implicit tuple sets for each (category, style) pair. We first generate $M=30$ unique starting phrases for the tuple set using the \texttt{STARTING\_CONVERSATION\_PROMPT}, so that each reply begins differently (Figure~\ref{fig:12_conv_gen_arith_multi} for the Arithmetic, Figure~\ref{fig:13_conv_gen_semantic_multi} for the  World-Knowledge, and Figure~\ref{fig:14_conv_gen_temporal_multi} for the Temporal reasoning). Then, for each tuple, we provide the forum topic, its base question, one starting phrases, and the tuple data to the \texttt{CONVERSATION\_GENERATION\_PROMPT} configured for forum responses; category-specific templates and requirements again appear in Figure~\ref{fig:12_conv_gen_arith_uni} for the Arithmetic, Figure~\ref{fig:13_conv_gen_semantic_multi} for the World-Knowledge, and Figure~\ref{fig:14_conv_gen_temporal_uni} for the Temporal reasoning. We generate exactly five sentences per response. Finally, each forum reply is passed to a separate LLM via the \texttt{FEATURE\_EXTRACTION\_PROMPT} to extract the original tuple, ensuring the response accurately encodes the tuple’s information (Figure~\ref{fig:12_feature_extract_arith_multi} for the Arithmetic, Figure~\ref{fig:13_feature_extract_semantic_multi} for the World-Knowledge, and Figure~\ref{fig:14_feature_extract_temporal_multi} for the Temporal reasoning).

\subsection{Human Evaluation Details}
\label{sec:app_human_eval}
We performed a small, stratified sanity check to complement automatic validation. For each reasoning category $\times$ discourse style cell, we randomly selected two queries (12 instances per cell). For each selected query, candidate passages were ranked with the \textsc{ReasonIR} retriever, and six passages were drawn by uniform sampling from two rank strata: top–20 (3) and bottom–20 (3). This yielded 72 passages in total. We assessed each passage for (i) \emph{fluency}, (ii) \emph{implicit support} of their corresponding queried fact (entailed but not stated verbatim), and (iii) \emph{absence of explicit leakage}. In this sample, all passages were fluent and passed the implicitness and non-leakage checks under our rubric.

\begin{table*}[t]
\centering
\renewcommand{\arraystretch}{1.15}
\resizebox{\linewidth}{!}{%
\begin{tabular}{l|c!{\vrule width .4pt}c!{\vrule width .4pt}c!{\vrule width .4pt}c}
\toprule
\multirow{2}{*}{\textbf{Reasoning}} &
\multicolumn{2}{c!{\vrule width .4pt}}{\textbf{Multi-speaker}} &
\multicolumn{2}{c}{\textbf{Uni-speaker}} 
\\ \cline{2-3} \cline{4-5}
& \textbf{\small{Each Pool}} & \textbf{\small{Each Document}} & \textbf{\small{Each Pool}} & \textbf{\small{Each Document}}\\ 
\midrule
\multirow{2}{*}{Arithmetic}     
& \multicolumn{1}{l|}{Topic,}
& \multicolumn{1}{l|}{Conversant, Brand, Model Years (Two random numbers}
& \multicolumn{1}{l|}{Main Conversant }
& \multicolumn{1}{l}{Second Conversant, Shopping} \\

& \multicolumn{1}{l|}{Forum-Base-Question}
& \multicolumn{1}{l|}{between 2013 up to 2024, the lower number is }
& 
& \multicolumn{1}{l}{item, Low priced brand } \\

&
& \multicolumn{1}{l|}{the price of the lower-priced item)} 
& 
& \multicolumn{1}{l}{name, High priced brand} \\
& \multicolumn{1}{l|}{}  & & &
\\
\multirow{2}{*}{World-Knowledge}
& \multicolumn{1}{l|}{Topic,} 
& \multicolumn{1}{l|}{Conversant,}
& \multicolumn{1}{l|}{Main Conversant }
& \multicolumn{1}{l}{Second Conversant, Landmark,}\\ 

& \multicolumn{1}{l|}{Forum-Base-Question} 
& \multicolumn{1}{l|}{Landmark}
&
& \multicolumn{1}{l}{Message Date }\\ 
& \multicolumn{1}{l|}{}  & & &
\\
\multirow{2}{*}{Temporal}
& \multicolumn{1}{l|}{Topic,  }
& \multicolumn{1}{l|}{Conversant, Item related to the Forum Base Question,} 
& \multicolumn{1}{l|}{Main Conversant }
& \multicolumn{1}{l}{Second Conversant, }\\ 

& \multicolumn{1}{l|}{Forum-Base-Question }
&  
& 
& \multicolumn{1}{l}{Daily work verb}\\ 
\bottomrule
\end{tabular}
}
\caption{%
\textbf{Entity-assignment granularity for each reasoning category and document style.} %
Items listed under \emph{Each Pool} are unique in that pool and are never reused across other pools. %
Items listed under \emph{Each Document} are unique within their own set and never reused across other Documents in that pool. %
All entities are drawn from the auxiliary lexical resources described in Section~\ref{sec:app_synthesizing_the_context}.%
}
\label{tab:tracks-styles-assign}
\end{table*}

\section{Results}
\label{sec:app_results}
\begin{table}[t]
\centering
\small
\resizebox{\columnwidth}{!}{%
\begin{tabular}{lcccc}
\toprule
\multirow{2}{*}{\textbf{Retriever}} 
& \multicolumn{3}{c}{\textbf{Reasoning}} & \multirow{2}{*}{\textbf{Average}} \\
\cmidrule(lr){2-4}
& \textbf{W. Know.}
& \textbf{Arithmetic}
& \textbf{Temporal}
& \\
\midrule
\multicolumn{5}{c}{\cellcolor{gray!15} Sparse Baseline}\\
\addlinespace
{BM25}      
& 9.55 & 7.42 & 6.83 & 7.93 \\
\specialrule{0.4pt}{\aboverulesep}{\belowrulesep}

\multicolumn{5}{c}{\cellcolor{gray!15} Late-Interaction}\\
\addlinespace
{ColBERT v2 }    
& 10.59 & \textbf{9.63} & 7.55 & 9.26 \\
\specialrule{0.4pt}{\aboverulesep}{\belowrulesep}

\multicolumn{5}{c}{\cellcolor{gray!15} Dense Encoders}\\
\addlinespace
{Contriever} 
& 11.19 & 8.84 & 8.48 & 9.50 \\
{Dragon+} 
& 11.97 & 9.47 & 8.26 & \textbf{9.90} \\
{ReasonIR}  
& \textbf{14.23} & 7.13 & 7.78 & 9.71 \\
\specialrule{0.4pt}{\aboverulesep}{\belowrulesep}

\multicolumn{5}{c}{\cellcolor{gray!15} Knowledge-Graph–Augmented Indexer}\\
\addlinespace
{HippoRAG 2} 
& 11.30 & 9.28 & \textbf{8.57} & 9.72  \\
\bottomrule
\end{tabular}%
}
\caption{\textbf{MRR@10} ranking metric scores for our reasoning category of World-Knowledge (W. Know.), Arithmetic, and Temporal, averaged over both Uni-speaker and Multi-speaker documents. The final “Average” column reports the mean MRR@10 across all reasoning categories.}
\label{tab:mrr10_avg}
\end{table}

\begin{figure*}[t]
\centering
\includegraphics[width=1.0\textwidth]{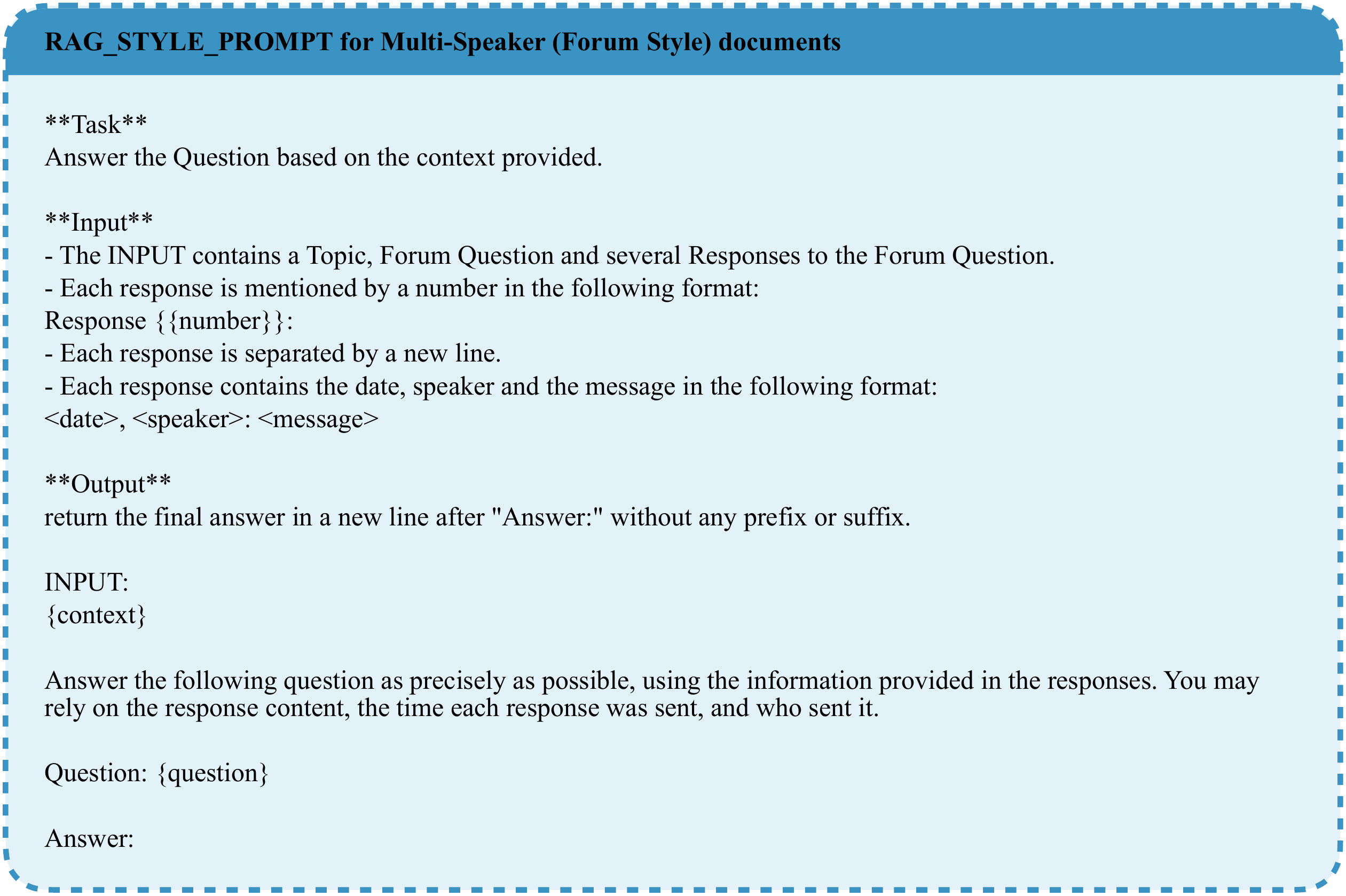}
\caption{Prompt for RAG-style experiment, while the input is forced to contain the positive document, in Multi-Speaker (Forum Style). This prompt is used for all the reasoning categories of the Arithmetic, World-Knowledge, and Temporal.}
\label{fig:15_LC_multi}
\end{figure*}

\begin{figure*}[t]
    \centering
    \includegraphics[width=1.0\textwidth]{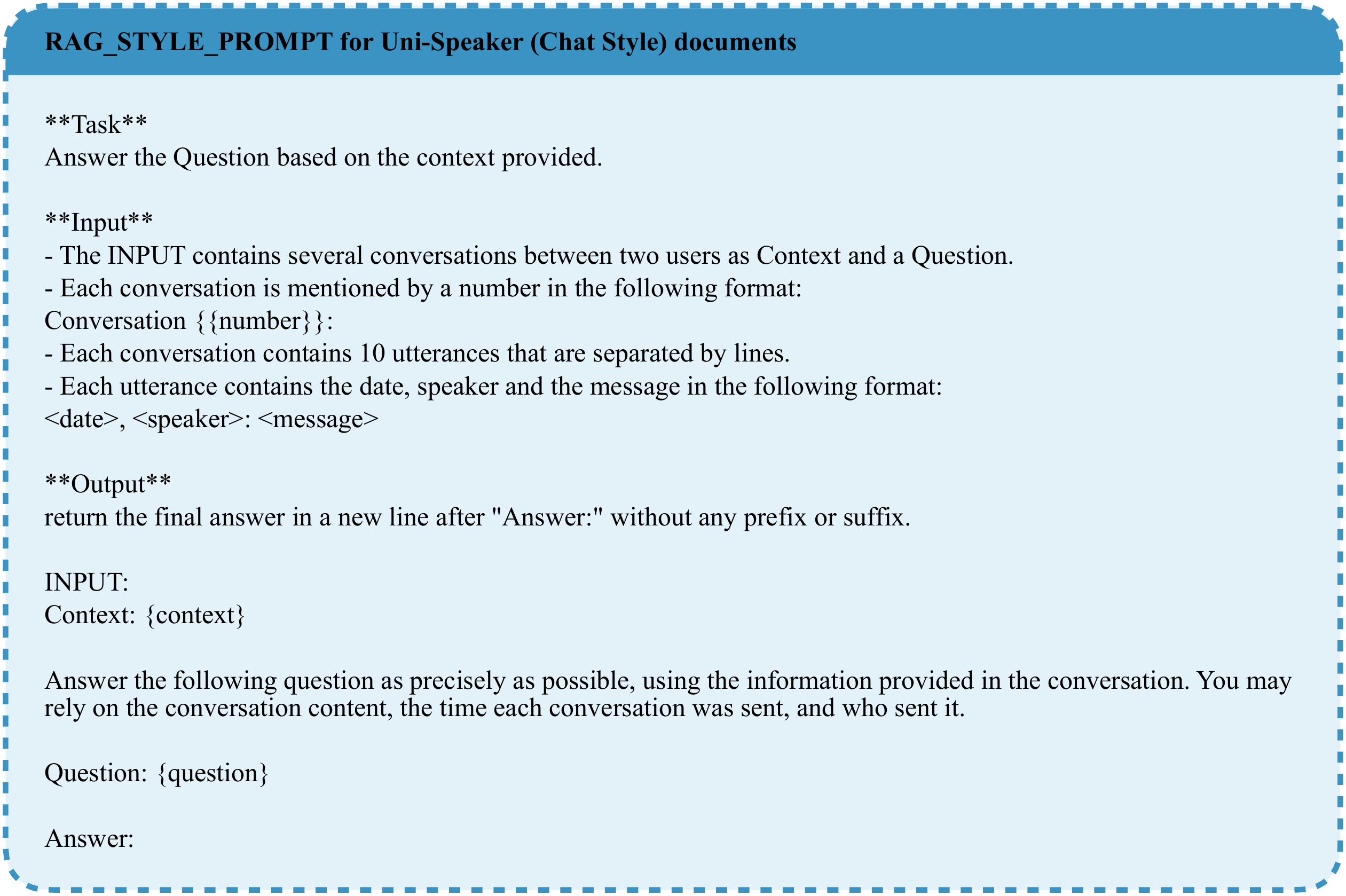}
    \caption{Prompt for RAG-style experiment, while the input is forced to contain the positive document, in Uni-Speaker (Chat Style). This prompt is used for all the reasoning categories of the Arithmetic, World-Knowledge, and Temporal.}
    \label{fig:15_LC_uni}
\end{figure*}

As explained in Section~\ref{sec:results}, our main retrieval metric is nDCG@10.  
For completeness, we also compute MRR@10, summarized in Table~\ref{tab:mrr10_avg}.  
The lower MRR@10 scores confirm that, across reasoning categories, the systems often fail to rank the positive documents first, underscoring the modest overall performance already suggested by nDCG.  
Granular results for the \emph{Uni-Speaker} and \emph{Multi-Speaker} settings are provided in Table~\ref{tab:app_full_metrics}.  
For the RAG-style evaluation, model outputs were generated using the prompt templates shown in Figures~\ref{fig:15_LC_multi} and~\ref{fig:15_LC_uni}, and then evaluated using ROUGE-1 Recall against the reference answer.

\begin{table*}[t]
\centering
\scriptsize
\resizebox{\linewidth}{!}{%
\begin{tabular}{l*{16}{c}}
\toprule
\multirow{4}{*}{\textbf{Experiment}}
& \multicolumn{4}{c}{\textbf{World-Knowledge}}
& \multicolumn{4}{c}{\textbf{Arithmetic}}
& \multicolumn{4}{c}{\textbf{Temporal}}
& \multicolumn{4}{c}{\textbf{Average}}\\[-0.2em]
\cmidrule(lr){2-5}\cmidrule(lr){6-9}\cmidrule(lr){10-13}\cmidrule(lr){14-17}
& \multicolumn{2}{c}{\textbf{Uni-speaker}}
& \multicolumn{2}{c}{\textbf{Multi-speaker}}
& \multicolumn{2}{c}{\textbf{Uni-speaker}}
& \multicolumn{2}{c}{\textbf{Multi-speaker}}
& \multicolumn{2}{c}{\textbf{Uni-speaker}}
& \multicolumn{2}{c}{\textbf{Multi-speaker}}
& \multicolumn{2}{c}{\textbf{Uni-speaker}}
& \multicolumn{2}{c}{\textbf{Multi-Speaker}}\\[-0.2em]
\cmidrule(lr){2-3}\cmidrule(lr){4-5}\cmidrule(lr){6-7}\cmidrule(lr){8-9}
\cmidrule(lr){10-11}\cmidrule(lr){12-13}\cmidrule(lr){14-15}\cmidrule(lr){16-17}
& \textbf{MRR@10} & \textbf{nDCG@10}
& \textbf{MRR@10} & \textbf{nDCG@10}
& \textbf{MRR@10} & \textbf{nDCG@10}
& \textbf{MRR@10} & \textbf{nDCG@10}
& \textbf{MRR@10} & \textbf{nDCG@10}
& \textbf{MRR@10} & \textbf{nDCG@10}
& \textbf{MRR@10} & \textbf{nDCG@10}
& \textbf{MRR@10} & \textbf{nDCG@10}\\
\midrule
\multicolumn{17}{c}{\cellcolor{gray!15}\textbf{Sparse Baseline}}\\
\addlinespace
BM25
& 9.23 & 14.18 & 9.86 & 15.20 & 9.23 & 14.05 & 5.61 & 8.07
& 6.33 & 10.20 & 7.33 & 11.76
& 8.26 & 12.81 & 7.60 & 11.68\\
\specialrule{0.4pt}{\aboverulesep}{\belowrulesep}

\multicolumn{17}{c}{\cellcolor{gray!15}\textbf{Late-Interaction}}\\
\addlinespace
ColBERT v2
& 9.57 & 14.54 & 11.60 & 17.03
& \textbf{9.45} & \textbf{14.62} & 9.81 & 15.30
& 6.85 & 10.86 & 8.25 & 13.12
& 8.62 & 13.34 & 9.89 & 15.15\\
\specialrule{0.4pt}{\aboverulesep}{\belowrulesep}

\multicolumn{17}{c}{\cellcolor{gray!15}\textbf{Dense Encoders}}\\
Contriever
& 9.77 & 14.77 & 12.60 & 18.23
& 8.23 & 12.87 & 9.45 & 14.52
& 8.17 & 12.22 & 8.78 & 13.25
& 8.72 & 13.29 & 10.28 & 15.33\\
\addlinespace
Dragon+
& 9.94 & 14.85 & 14.00 & 20.07
& 9.11 & 14.11 & 9.83 & 15.12
& 7.20 & 11.16 & 9.32 & 14.16
& 8.75 & 13.37 & 11.05 & 16.45\\
\addlinespace
ReasonIR
& 7.75 & 10.54 & \textbf{20.71} & \textbf{27.21}
& 3.87 &  5.68 & \textbf{10.39} & \textbf{15.89}
& 2.95 &  4.17 & \textbf{12.60} & \textbf{18.33}
& 4.86 &  6.80 & \textbf{14.57} & \textbf{20.48}\\
\specialrule{0.4pt}{\aboverulesep}{\belowrulesep}

\multicolumn{17}{c}{\cellcolor{gray!15}\textbf{Knowledge-Graph–Augmented Indexer}}\\
\addlinespace
HippoRAG 2
& \textbf{9.95} & \textbf{14.94} & 12.66 & 18.30
& 8.31 & 12.97 & 10.26 & 15.29
& \textbf{8.18} & \textbf{12.24} & 8.96 & 13.42
& \textbf{8.81} & \textbf{13.38} & 10.63 & 15.67\\
\bottomrule
\end{tabular}}
\caption{\textbf{MRR@10} and \textbf{nDCG@10} for each reasoning category and discourse setting.  
The maximum value in every metric column is bold-faced.  The final “Average” block shows per-setting means over the three categories.}
\label{tab:app_full_metrics}
\end{table*}

\begin{figure*}[t]
    \centering
    \includegraphics[width=\textwidth]{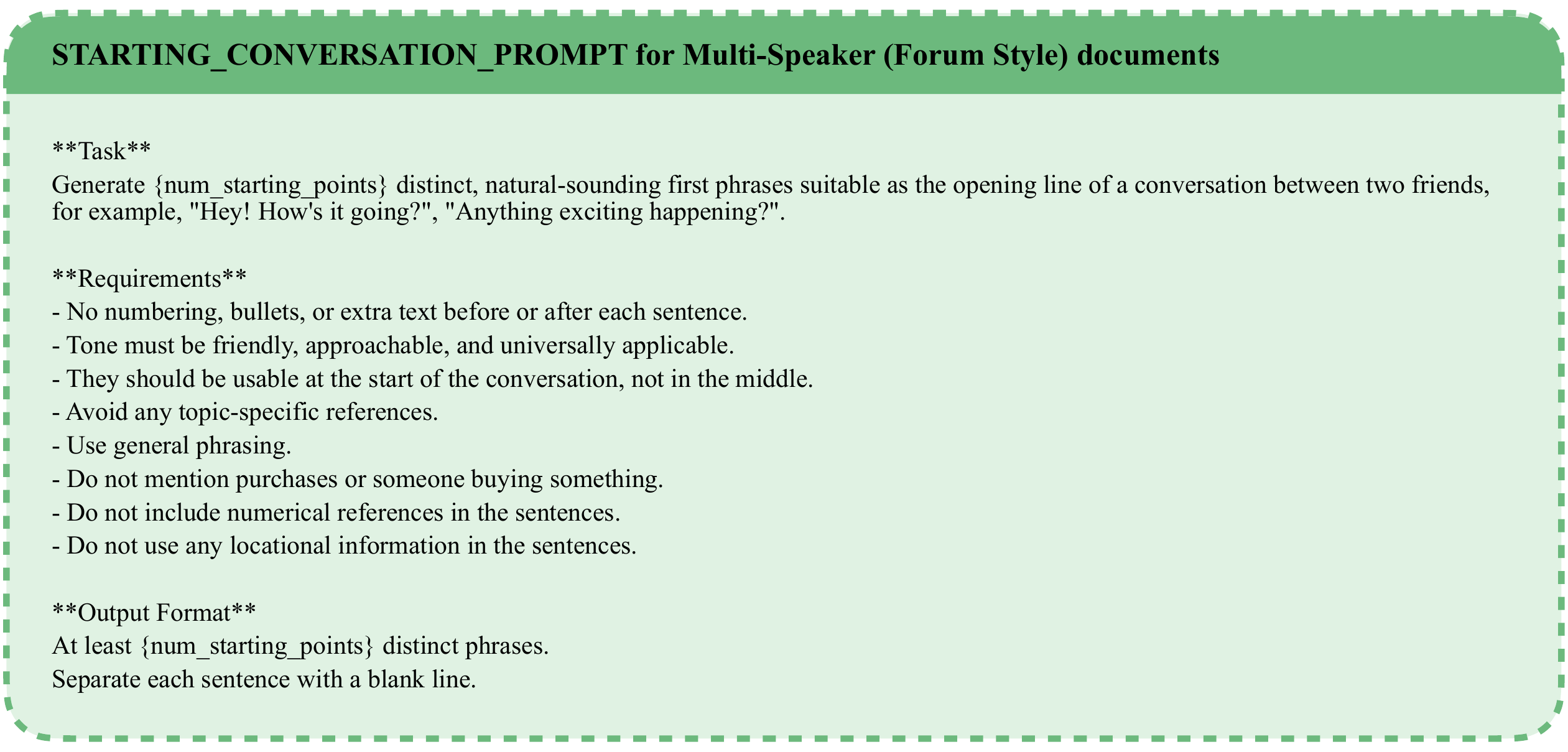}
    \caption{Prompt for generating a list of the first phrase in Multi-Speaker (Forum Style) documents. This prompt is used for all the reasoning categories of the Arithmetic, World-Knowledge, and Temporal.}
    \label{fig:11_Starting_conv_multi}
\end{figure*}





\end{document}